\documentclass[lettersize,journal]{IEEEtran}
\usepackage{amsmath,amsfonts}
\usepackage{algorithmic}
\usepackage{algorithm}
\usepackage{array}
\usepackage[caption=false,font=normalsize,labelfont=sf,textfont=sf]{subfig}
\usepackage{textcomp}
\usepackage{stfloats}
\usepackage{url}
\usepackage{verbatim}
\usepackage{graphicx}
\usepackage{cite}
\usepackage{booktabs}
\usepackage{multirow}
\usepackage{tabularx}
\usepackage{pifont}
\usepackage[utf8]{inputenc}
\usepackage[T1]{fontenc}
\usepackage[table]{xcolor}
\usepackage{booktabs}
\usepackage{multirow}

\newcommand{\cmark}{\ding{51}}

\begin{document}

\title{BioKD: Selective Physiology-to-Video Knowledge Distillation via Reliability Gate for Emotion Recognition}

%\author{IEEE Publication Technology,~\IEEEmembership{Staff,~IEEE,}
        % <-this % stops a space
%\thanks{This paper was produced by the IEEE Publication Technology Group. They are in Piscataway, NJ.}% <-this % stops a space
%\thanks{Manuscript received April 19, 2021; revised August 16, 2021.}}

\author{
Bojing Hou,
Ruohao Li,
Yitong Zhu,
Hongjun Liu,
Luwen Yu,
and Yuyang Wang%
\thanks{Bojing Hou, Ruohao Li, Yitong Zhu, Luwen Yu, and Yuyang Wang are with The Hong Kong University of Science and Technology (Guangzhou), Guangzhou, China (e-mail: \{bhou870, rli777, yzhu162\}@connect.hkust-gz.edu.cn; luwenyu@hkust-gz.edu.cn; yuyangwang@hkust-gz.edu.cn).}%
\thanks{Hongjun Liu is with New York University, New York, NY, USA (e-mail: janne150.lhj@gmail.com).}%
\thanks{Corresponding author: Yuyang Wang}
}


\markboth{IEEE Transactions on Affective Computing}%
{Hou \MakeLowercase{\textit{et al.}}: BioKD: Selective Physiology-to-Video Knowledge Distillation}


\maketitle

\begin{abstract}
To address the limitations of video-based emotion recognition under ambiguous or socially masked behavioral cues, as well as the poor deployability of physiological signals, this paper proposes a reliability-aware physiology-to-video knowledge distillation framework, termed BioKD. The proposed framework leverages physiological signals as privileged information during training to guide a video-based student model in learning deep affective representations, while relying solely on non-intrusive video inputs at inference time. To cope with the high noise and instability of physiological teacher supervision caused by inter-subject variability, signal artifacts, and temporal inconsistency, BioKD incorporates a sample-wise reliability-aware gating mechanism together with a progressive distillation strategy. By adaptively regulating the strength of knowledge transfer, the framework suppresses negative transfer induced by unreliable physiological supervision and enables more stable cross-modal distillation. Experiments on DEAP and AMIGOS show that BioKD consistently outperforms representative baselines under both trial-wise and subject-wise evaluation protocols for valence and arousal recognition. For example, BioKD achieves 68.01\% on DEAP (trial-wise arousal) and 65.29\% under the more challenging subject-wise setting, demonstrating improved performance under a subject-independent evaluation setting. Further analyses show that BioKD effectively mitigates overconfident teacher errors and outperforms an entropy-only weighting strategy, confirming the importance of explicitly modeling supervision reliability. 
In addition, BioKD introduces no additional inference-time overhead relative to the same video student architecture and removes the need for physiological sensing and multimodal synchronization. 
\end{abstract}


\section{Introduction}

Emotion recognition is a core problem in affective computing, with broad applications in human--computer interaction, mental health assessment, and intelligent content understanding~\cite{poria2017review,dzedzickis2020human,ahmad2022survey}. In recent years, substantial progress has been achieved by learning emotion representations from observable behavioral cues (e.g., facial expressions and body movements) benefitting from their accessible data acquisition and efficient deployment~\cite{li2020deep,ahmed2019emotion,li2025multimodal,pan2023multimodal,udahemuka2024multimodal}.  However, these approaches implicitly assume that observable behaviors can faithfully reflect an individual’s internal affective state~\cite{gross1993emotional}, an assumption frequently violated by emotion regulation and social masking, leading to systematic discrepancies between external behaviors and genuine emotional experiences~\cite{pantic2003toward}.

To bridge the gap between observable behavioral cues and individuals’ underlying internal affective states, physiological signals, such as electroencephalography (EEG), electrodermal activity (EDA), and blood volume pulse (BVP), have been widely explored as complementary sources of affective information, as they capture central and autonomic nervous system activity and provide internal affective cues that complement observable behavior~\cite{koelstra2011deap,critchley2013interaction,zhu2025hierarchical,lopez2024phemonet}. 
Despite their high affective informativeness, physiological signals are inherently noisy, temporally misaligned with emotion labels, and costly or intrusive to acquire, which severely limits their reliability and feasibility as inference-time inputs~\cite{lotte2018review,liu2019generalized,schmidt2018introducing}.
Collectively, these factors render physiological signals a rich yet unreliable modality when used directly as inference-time inputs. 
Accordingly, how physiological signals should be effectively utilized in emotion recognition models warrants careful reconsideration. Conventional multimodal fusion approaches typically incorporate both physiological and behavioral signals at inference time, inevitably inheriting the aforementioned limitations of physiological inputs~\cite{ramaswamy2024multimodal,strizhkova2024mvp,yin2017recognition}. 
Recent studies have explored using information-rich but deployment-unfriendly modalities exclusively during training to improve the robustness of primary modalities~\cite{vapnik2015learning,zhu2025real,aslam2023privileged,hoffman2016learning,jemiolo2022datasets}. Inspired by this paradigm, we treat physiological signals as training-only supervision, while relying solely on video at inference time to ensure accessibility and deployability.

Knowledge distillation (KD) provides a natural mechanism to realize this paradigm, enabling the transfer of supervision from a physiology-based teacher to a video-based student model~\cite{hinton2015distilling}. 
However, when the teacher model itself is built upon unreliable physiological signals, the quality and reliability of its supervisory signals can vary substantially across samples due to noise, instability, and pronounced inter-individual variability.
Under such conditions, applying na\"{\i}ve knowledge distillation in a uniform manner may not only fail to improve student performance, but can even introduce misleading supervision and cause negative transfer~\cite{mirzadeh2020improved,muller2019does,grathwohl2019your}. 

Consequently, the main challenge addressed in this work is how to safely distill affective knowledge from unreliable physiological teachers into deployable video-based models without introducing negative transfer. 
To address these challenges, we propose \textbf{BioKD}, a selective physiology-to-video knowledge distillation framework that adopts a teacher--student paradigm, where a physiology-only teacher provides complementary affective supervision during training.
%In doing so, BioKD guides the student to learn more discriminative emotion representations without altering inference-time inputs or deployment conditions. 
The core of BioKD is a \emph{sample-wise reliability gating mechanism} that models variations in the reliability of physiological teacher supervision and selectively regulates knowledge transfer, distilling informative cues when predictions are stable while suppressing unreliable signals to mitigate negative transfer.
%The core of BioKD lies in a \emph{sample-wise reliability gating mechanism} that explicitly models the reliability variation of physiological teacher supervision across samples and selectively regulates the knowledge transfer process of teacher supervision to the distillation loss. When teacher predictions are relatively stable, informative implicit affective cues are effectively distilled; conversely, when supervision is noisy or uncertain, unreliable signals are suppressed to mitigate negative transfer. 
Furthermore, because the video student has not yet learned stable representations during early training, BioKD incorporates a \emph{progressive distillation strategy} that gradually introduces and strengthens teacher supervision over training stages, enabling reliability modeling to take effect in a smoother and more stable manner. The overall framework consists of physiological teacher modeling, sample-wise reliability assessment, and selective distillation training, with the training workflow and module interactions illustrated in Fig.~\ref{fig:framework}. 

Extensive experiments on DEAP and AMIGOS show that BioKD consistently outperforms representative distillation baselines, demonstrating consistent improvements under both trial-wise and subject-wise evaluation protocols, highlighting the practical effectiveness of reliability-aware physiology-to-video distillation. Overall, the main contributions of this work are summarized as follows:
%Concretely, BioKD improves video-based emotion recognition accuracy by 3\% on DEAP and 4\% on AMIGOS compared to strong student-only baselines, demonstrating the practical effectiveness of reliability-aware distillation.

\begin{itemize}
    \item We propose \textbf{BioKD}, a selective physiology-to-video knowledge distillation framework that repositions physiological signals as a \emph{training-only knowledge source}, improving video-based emotion recognition performance without modifying inference-time inputs or deployment conditions.
    \item We design a \emph{sample-wise reliability gating mechanism} that explicitly models the reliability variation of physiological teacher supervision, enabling selective knowledge transfer and mitigating negative transfer caused by noisy and unstable physiological signals.
    \item We systematically integrate a \emph{progressive distillation strategy} into the proposed framework to stabilize optimization when distilling from unreliable physiological teachers, facilitating more effective reliability-aware knowledge transfer.
%    \item Extensive experiments on multiple benchmark emotion recognition datasets demonstrate that BioKD consistently and significantly outperforms na\"{\i}ve distillation methods and student-only baselines, validating both the effectiveness and necessity of the proposed framework. 
\end{itemize}

\section{Related Works}

\subsection{Knowledge Distillation and Cross-Modal Distillation}

Knowledge distillation (KD) is a widely studied paradigm for transferring knowledge from high-capacity teacher models to lightweight students, enabling competitive performance with reduced model complexity and inference cost across various domains~\cite{hinton2015distilling,gou2021knowledge,mansourian2025comprehensive,wu2025comprehensive,pan2023review}.

Existing KD methods mainly differ in the form of knowledge transferred from the teacher to the student. Logit-based distillation aligns output probability distributions to convey inter-class relationships~\cite{hinton2015distilling}, while feature-based distillation aligns intermediate representations to transfer richer structural information~\cite{romero2014fitnets}. Relational distillation, multi-layer distillation, and teacher-assistant strategies have further been proposed to transfer sample relationships, integrate supervision across representation levels, and reduce teacher--student capacity gaps~\cite{park2019relational,zagoruyko2016paying,mirzadeh2020improved}. Despite their success, most KD methods assume that teacher supervision is consistently reliable and therefore apply distillation objectives uniformly. This assumption overlooks cases in which teacher predictions or representations are informative for some samples but inaccurate or misleading for others, potentially introducing negative transfer.

With the development of multimodal learning, KD has been extended to cross-modal settings, where information-rich but deployment-unfriendly modalities supervise unimodal or lightweight student models~\cite{gupta2016cross,chen2024vision,jia2025cross}. This paradigm reduces inference-time sensing and computational costs while preserving complementary information available during training. Compared with same-modality distillation, cross-modal KD must additionally address substantial differences in input structure, temporal characteristics, and representation distributions between teacher and student modalities. Consequently, direct feature alignment may be unreliable when heterogeneous modalities encode different aspects of the same underlying semantic state.

In affective computing, existing cross-modal distillation methods typically align predictions or intermediate features from multimodal or physiological teachers with unimodal students~\cite{liu2023emotionkd,ma2023transformer,hussain2025optimised}. However, they largely focus on what knowledge should be transferred or how heterogeneous representations should be aligned, while paying less attention to whether teacher supervision is trustworthy for each individual sample. This limitation is particularly important when the teacher itself relies on noisy and highly variable physiological measurements.

\subsection{Reliability Challenges in Physiological Knowledge Distillation}

When teacher models are built upon physiological signals or other high-noise modalities, their predictions may exhibit substantial sample-level variability~\cite{lawhern2018eegnet,makantasis2023lab}. Physiological signals are susceptible to acquisition noise, temporal misalignment, sensor artifacts, and inter-subject variability, which can produce unstable or even misleading teacher predictions, including overconfident errors~\cite{guo2017calibration,bota2019review}. Therefore, the average predictive performance of a physiological teacher does not imply that its supervision is equally reliable for every sample. A teacher may provide useful complementary information for one sample while producing confident but incorrect guidance for another.

Under such conditions, na\"{\i}ve distillation may propagate noisy supervision into the student and cause negative transfer~\cite{zhang2019your,lai2024online}. Uniform distillation is particularly problematic because it assigns similar importance to reliable and unreliable teacher outputs, allowing erroneous supervision to influence student optimization. Although uncertainty modeling, confidence-based weighting, and robust loss functions have been explored to mitigate distillation noise~\cite{yang2024uncertainty,saganowski2022emotion}, these approaches typically treat prediction uncertainty as a proxy for supervision quality. However, confidence and correctness are not equivalent: a low-confidence prediction may still contain useful information, whereas an overconfident prediction may remain incorrect.

Under non-stationary physiological noise, this limitation becomes more pronounced because teacher reliability may vary across samples and training stages. Global distillation coefficients or coarse-grained weighting strategies cannot adequately capture these dynamic variations, while decisions based solely on instantaneous confidence may respond strongly to transient fluctuations. Reliable physiology-to-video distillation therefore requires a sample-wise mechanism that considers not only the teacher's current prediction strength but also the historical consistency of teacher--student representations and predictions.

Motivated by these limitations, we study a practical cross-modal distillation setting in which the teacher relies on informative yet potentially unreliable physiological signals. Unlike prior approaches that assume uniformly reliable supervision, we explicitly model sample-wise reliability and selectively regulate the strength of knowledge transfer. By treating physiological signals as training-only privileged information and combining reliability-aware gating with progressive distillation, BioKD aims to preserve useful internal affective cues while suppressing unreliable physiological guidance, enabling more robust cross-modal affective knowledge distillation and video-only inference.

\section{Methodology}

\subsection{The Core Challenge: Reliability Gap in Physiological Teachers}
\label{sec:reliability_gap}

\paragraph{Failure of Conventional Assumptions.}
Given a multimodal dataset $\mathcal{D}=\{(x_i^v,x_i^p,y_i)\}_{i=1}^{N}$, standard knowledge distillation typically assumes that the prediction outputs of the teacher model provide reliable supervision for the student. However, physiological signals are susceptible to non-stationary noise, including motion artifacts, unstable sensor contact, and inter-subject variability.

Let $\mathbf{s}_i^T=\mathcal{T}(x_i^p)$ denote the prediction logits produced by the physiological teacher. We use the following output-level abstraction:
\begin{equation}
\mathbf{s}_i^T = \mathbf{s}_i^\star + \boldsymbol{\epsilon}_i,
\label{eq:noise}
\end{equation}
where $\mathbf{s}_i^\star$ denotes the latent clean supervision that would be obtained under an ideal acquisition condition, and $\boldsymbol{\epsilon}_i$ summarizes sample-dependent distortions propagated from physiological artifacts to the teacher output. Equation~\eqref{eq:noise} does not assume that noise is physically added to the logits; it provides an output-level abstraction of the unreliable teacher supervision observed by the student.

\begin{figure}[t]
    \centering
    \includegraphics[width=0.9\linewidth]{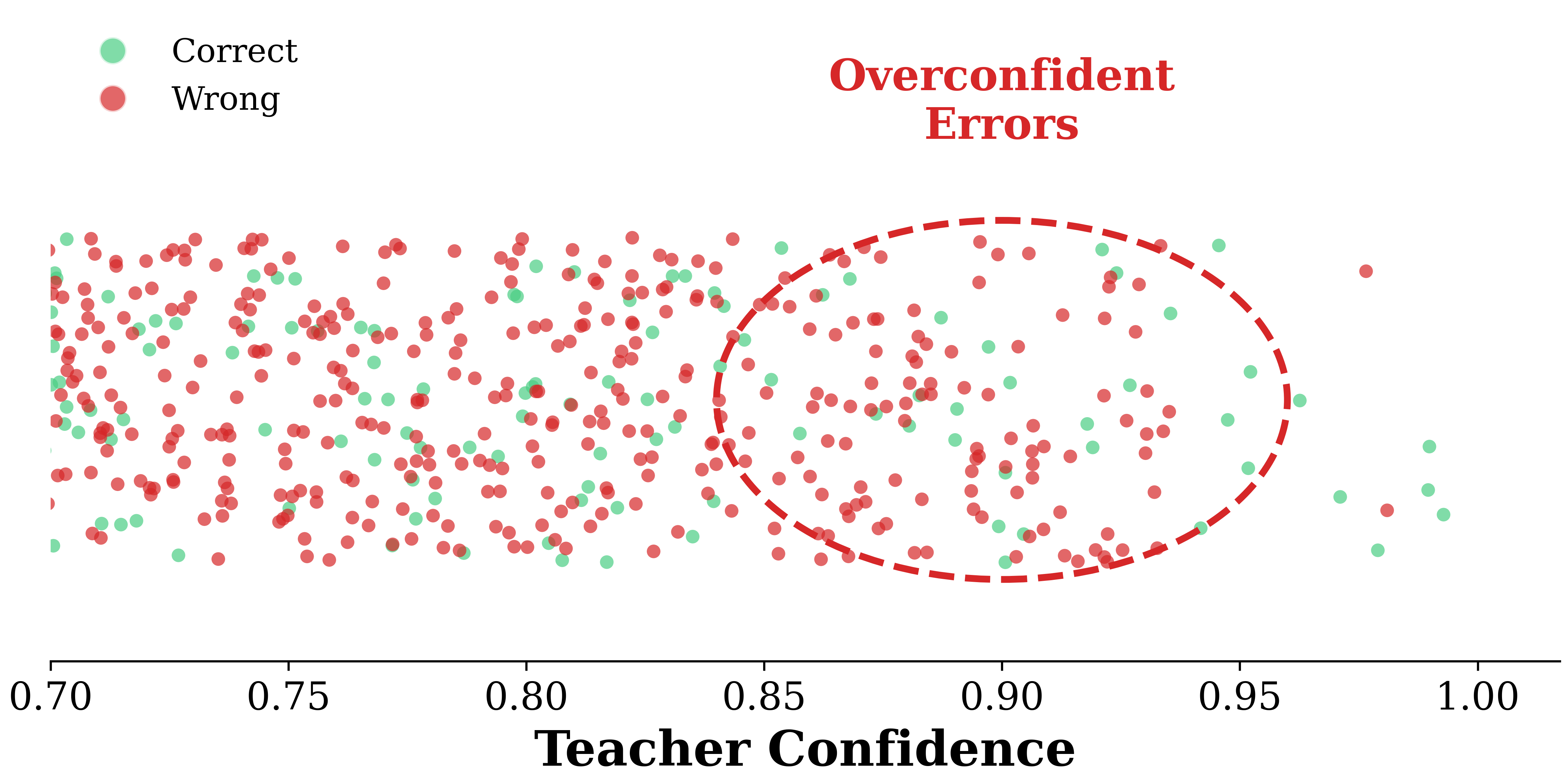}
    \caption{\textbf{Confidence–accuracy reliability analysis of the physiological teacher, highlighting overconfident errors.} 
 }
    \label{fig:reliability_gap} 
\end{figure}

\paragraph{Overconfident Errors.}
To validate the hypothesis that physiological noise $\epsilon_i$ distorts the teacher's outputs (as formulated in Eq.~\ref{eq:noise}), we analyze the relationship between the teacher's prediction confidence ($\operatorname{conf}_i^T$) and its actual correctness on the validation set. As shown in Fig.~\ref{fig:reliability_gap}, the physiological teacher exhibits a clear \textbf{confidence--accuracy mismatch}: even in the high-confidence regime (e.g., $\operatorname{conf}_i^T > 0.85$), a noticeable number of samples are still misclassified (highlighted by the red dashed circle). This indicates that high teacher confidence does not necessarily imply trustworthy supervision~\cite{guo2017calibration}.

Beyond overconfidence, the reliability of physiological supervision may fluctuate under non-stationary conditions. To address this issue, our framework introduces temporal reliability accumulation, as detailed in Section~\ref{sec:reliability_gate}.

\paragraph{Risk of Negative Transfer.}
These observations indicate a potential risk: ignoring sample-wise reliability variations and directly applying a na\"{\i}ve distillation loss may encourage the student to imitate distorted or misleading teacher supervision, as summarized by the output-level distortion term $\boldsymbol{\epsilon}_i$ in Eq.~\eqref{eq:noise}. Because the reliability of physiological teacher outputs varies across samples and acquisition conditions, uniform distillation may propagate teacher errors to the student and result in negative transfer. This finding motivates a mechanism capable of tracking the historical behavior of teacher--student supervision and providing a stabilized reliability estimate for selective knowledge transfer.

\subsection{Overview of the BioKD Framework}
\label{sec:framework_overview}

\begin{figure*}[t]
    \centering
    \includegraphics[width=0.95\linewidth]{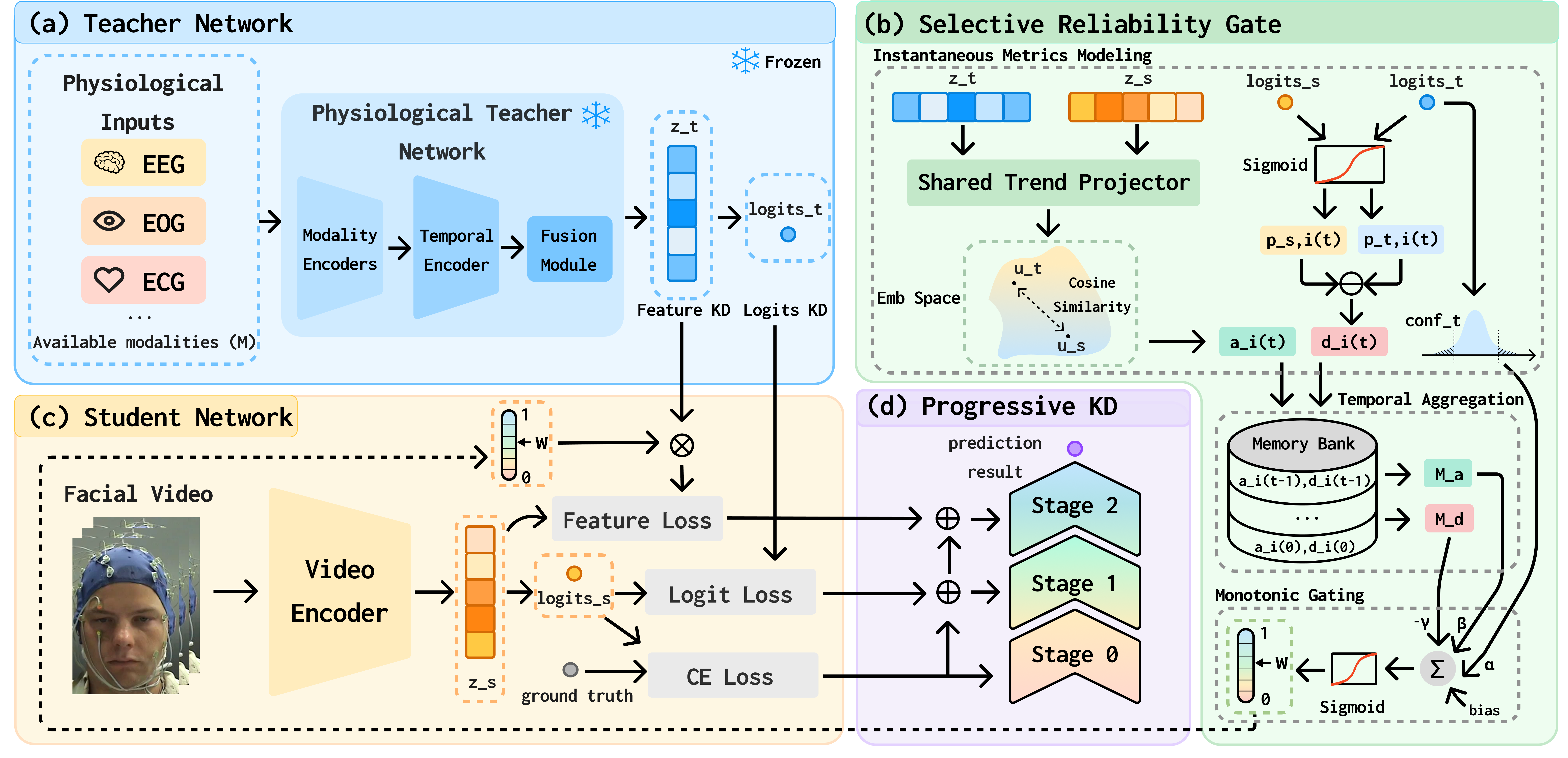}
    \caption{\textbf{Overview of the BioKD framework.}
    The framework performs physiology-to-video knowledge distillation via reliability-aware gating.
    (a) A pre-trained physiological teacher provides expert feature supervision.
    (b) A selective reliability gate dynamically generates the distillation weight $w$ using historical alignment information.
    (c) A video student learns emotion representations under teacher guidance.
    (d) A progressive distillation schedule gradually transitions from basic classification to gated feature alignment.}
    \label{fig:framework}
\end{figure*}

As outlined in Fig.~\ref{fig:framework}, BioKD performs physiology-to-video knowledge distillation in a reliability-aware, progressive manner. During training, a pre-trained physiological teacher provides complementary affective supervision, while a selective reliability gate regulates knowledge transfer based on the estimated reliability of teacher predictions. To stabilize training, a progressive distillation schedule gradually introduces gated supervision.

\subsection{Selective Reliability Gate}
\label{sec:reliability_gate}

To address reliability fluctuations in physiological teacher supervision, we design the \textbf{Selective Reliability Gate} as a closed-loop system that integrates \emph{instantaneous metric modeling}, \emph{temporal aggregation}, and \emph{monotonic gating}.
Its objective is to convert noisy alignment signals into robust historical reliability estimates and generate sample-wise, learnable distillation weights for controlled cross-modal feature transfer (as shown in Fig.~\ref{fig:framework}(b)).

\subsubsection{From Cross-Modal Alignment to Instantaneous Metrics.}
Due to the manifold discrepancy between the visual representation $\mathbf{z}_i^S$ extracted by the student and the physiological representation $\mathbf{z}_i^T$ extracted by the teacher~\cite{zhang2022visual}, direct comparison in their original feature spaces may be unreliable. To enable cross-modal comparison, we introduce a shared trend projector $\phi(\cdot)$ that maps both representations into a common semantic space using the same parameters:
\begin{equation}
\mathbf{u}_i^S
=
\phi(\mathbf{z}_i^S),
\qquad
\mathbf{u}_i^T
=
\phi(\mathbf{z}_i^T).
\label{eq:shared_projection}
\end{equation}
Here, $\mathbf{u}_i^S$ and $\mathbf{u}_i^T$ denote the projected student and teacher representations, respectively. Parameter sharing provides a consistent metric space for measuring cross-modal feature agreement.

Within this shared space, we define the \textbf{Feature Agreement} $a_i^{(t)}$ to quantify the instantaneous semantic alignment between the teacher and the student:
\begin{equation}
\begin{aligned}
a_i^{(t)} = \frac{1}{2}\left(
\frac{\mathbf{u}_i^S \cdot \mathbf{u}_i^T}
{\|\mathbf{u}_i^S\|_2 \, \|\mathbf{u}_i^T\|_2}
+ 1
\right),
\end{aligned}
\end{equation}
where the output is rescaled to $[0,1]$ for numerical stability.
This metric captures whether the teacher and student representations exhibit a consistent semantic trend at the current training step.

To capture teacher--student conflict at the decision level, we first convert the single-logit outputs into positive-class probabilities:
\begin{equation}
p_i^S=\sigma(s_i^S),
\qquad
p_i^T=\sigma(s_i^T),
\end{equation}
where $s_i^S$ and $s_i^T$ denote the student and teacher logits, respectively. The instantaneous prediction disagreement is then defined as
\begin{equation}
d_i^{(t)}
=
\left|
p_i^S-p_i^T
\right|.
\label{eq:prediction_disagreement}
\end{equation}

Because $p_i^T$ alone represents the positive-class probability rather than prediction confidence, the teacher confidence is defined symmetrically as
\begin{equation}
\operatorname{conf}_i^T
=
\max
\left(
p_i^T,1-p_i^T
\right).
\label{eq:teacher_confidence}
\end{equation}
This definition assigns high confidence to decisive predictions of either class while remaining independent of the predicted class identity.

Unlike $a_i^{(t)}$ and $d_i^{(t)}$, this signal does not describe the teacher--student relationship, but instead reflects the teacher’s own assessment of sample saliency or discriminability at the current training step.
Together, $\{a_i^{(t)}, d_i^{(t)}, \operatorname{conf}_i^T\}$ constitute three complementary instantaneous reliability cues observable at each iteration, where $a_i^{(t)}$ and $d_i^{(t)}$ describe relational quality, and $\operatorname{conf}_i^T$ captures sample saliency.

\subsubsection{Time-Scale Separation and Consistency Memory.}

Although these instantaneous metrics are informative, physiological artifacts introduce short-term, non-stationary fluctuations, making direct use of per-step signals unreliable for distillation.

More fundamentally, these signals differ in their temporal characteristics.
To validate this observation, we conduct a pilot temporal analysis on the validation set.
We find that while the instantaneous agreement $a_i^{(t)}$ and disagreement $d_i^{(t)}$ exhibit noticeable fluctuations, exponential smoothing effectively suppresses high-frequency variations.
Specifically, the standard deviation of temporal differences is reduced from $0.0356$ to $0.0049$ for $a_i^{(t)}$ and from $0.0662$ to $0.0070$ for $d_i^{(t)}$ after smoothing.
In contrast, the teacher confidence $\operatorname{conf}_i^T$ remains relatively stable at the instantaneous level ($\sigma_{\Delta}=0.0373$) and is thus treated as a per-iteration saliency cue.

Motivated by this observation, we perform an explicit \textbf{time-scale separation}.
Only the structurally meaningful signals $a_i^{(t)}$ and $d_i^{(t)}$ are treated as \emph{trustworthiness cues} and accumulated over time using a \emph{Consistency Memory Bank}, while $\operatorname{conf}_i^T$ is retained as an instantaneous \emph{saliency cue} and directly fed into the gating function.
Specifically, for each training sample $i$, the agreement and disagreement memories are updated independently:
\begin{equation}
\begin{aligned}
\mathcal{M}_{\mathrm{agree},i}^{(t)}
&=
m\mathcal{M}_{\mathrm{agree},i}^{(t-1)}
+
(1-m)a_i^{(t)},\\
\mathcal{M}_{\mathrm{disag},i}^{(t)}
&=
m\mathcal{M}_{\mathrm{disag},i}^{(t-1)}
+
(1-m)d_i^{(t)},
\end{aligned}
\label{eq:memory_update}
\end{equation}
where $t$ denotes the training iteration and $m\in[0,1)$ is the memory momentum. These updates yield smoothed historical estimates of feature agreement and prediction disagreement without concatenating the two reliability cues.
The memory thus yields robust historical estimates of feature agreement $\mathcal{M}_{\text{agree},i}$ and prediction disagreement $\mathcal{M}_{\text{disag},i}$.
The consistency memory is maintained exclusively for samples in the training split and is updated only during student optimization. Neither the memory bank nor the reliability gate is required during validation or test inference, and both are discarded after training. Consequently, validation and test samples are never used to update the historical reliability statistics or the parameters of the student model.

\subsubsection{Monotonic Trend Gating and Feedback Regulation.}
Given the stabilized historical statistics, the gating module combines the teacher confidence, historical feature agreement, and historical prediction disagreement into an unconstrained reliability score:
\begin{equation}
\begin{aligned}
r_i
={}&
b
+
\operatorname{Softplus}(\alpha)
\operatorname{conf}_i^T
+
\operatorname{Softplus}(\beta)
\mathcal{M}_{\mathrm{agree},i}
\\
&-
\operatorname{Softplus}(\gamma)
\mathcal{M}_{\mathrm{disag},i}.
\end{aligned}
\label{eq:gate_score}
\end{equation}
The final sample-wise reliability weight is obtained through
\begin{equation}
w_i
=
\max
\left\{
\epsilon,\,
\sigma(r_i)
\right\},
\label{eq:gate_weight}
\end{equation}
where $\epsilon$ denotes the minimum gate weight. The Softplus parameterization constrains the scaling coefficients to be non-negative, such that teacher confidence and historical agreement can only increase the gate weight, whereas persistent disagreement can only decrease it.

This formulation enforces an explicit \textbf{monotonicity constraint}: both sample saliency and historical agreement can only increase the distillation weight, whereas persistent disagreement always suppresses it.
In particular, the disagreement term is designed to identify and mitigate \emph{high-confidence traps}—cases where the teacher exhibits strong confidence but remains consistently misaligned with the student—thereby preventing the propagation of unreliable gradients.

The resulting weight $w_i$ is fed back along the dashed path shown in Fig.~\ref{fig:framework}(b) to modulate the feature distillation loss.
For samples with stable and trustworthy teacher supervision, the gate opens to encourage deep feature alignment; for samples affected by physiological noise or persistent conflicts, the gate attenuates gradient flow.

To prevent the overall gate activation from drifting toward a degenerate extreme, we constrain its mini-batch mean using a mean-anchoring regularizer:
\begin{equation}
\mathcal{L}_{\mathrm{gate}}
=
\left(
\frac{1}{B}
\sum_{i=1}^{B}
w_i
-
\rho
\right)^2,
\label{eq:gate_regularization}
\end{equation}
where $\rho=0.4$ denotes the target mean gate activation. This regularizer controls the global activation level while allowing the monotonic gate to retain sample-specific variations induced by the three reliability cues.

Together, the monotonic cue structure and mean-anchoring constraint enable sample-wise regulation of cross-modal feature transfer while preventing global gate collapse.

\subsection{Progressive Distillation Strategy}
\label{sec:progressive_kd}

Although the selective reliability gate can dynamically modulate the strength of cross-modal feature distillation, introducing deep feature alignment at the early stage of training may still expose the student to unstable physiological supervision. 
In physiology-to-video distillation, the student has not yet learned stable facial representations at the beginning of training, and enforcing complex distillation constraints too early may lead to negative transfer.

To address this issue, BioKD adopts a \textbf{progressive distillation strategy} that organizes training into three consecutive stages, as illustrated in Fig.~\ref{fig:framework}(d), allowing knowledge transfer to proceed in a coarse-to-fine manner.

\paragraph{Stage 0: Warm-up ($e \le E_0$).}
The student model is trained only with ground-truth supervision, without any teacher guidance, to establish a stable visual representation baseline.

\paragraph{Stage 1: Logits-level Alignment ($E_0 < e \le E_1$).}
The teacher's logits are introduced as soft supervision to perform semantic-level distillation.
This stage leverages the cross-modal robustness of logits to guide the student toward learning inter-class semantic relations.
Meanwhile, the memory bank in the reliability gate starts tracking teacher--student consistency patterns in the background.

\paragraph{Stage 2: Reliability-driven Feature Refinement ($e > E_1$).}
Feature-level distillation is activated under the control of the reliability gate.
The sample-wise distillation weight $w_i$ dynamically adjusts the strength of feature alignment, ensuring that deep alignment is emphasized only when the teacher is considered reliable.

\paragraph{Distillation Objectives.}
Let $s_i^S$ and $s_i^T$ denote the single-logit outputs of the student and teacher, respectively, and let $y_i\in\{0,1\}$ denote the binary emotion label. The task loss is implemented using binary cross-entropy with logits:
\begin{equation}
\mathcal{L}_{\mathrm{task}}
=
\frac{1}{B}
\sum_{i=1}^{B}
\operatorname{BCELogits}(s_i^S,y_i),
\end{equation}
where $B$ denotes the mini-batch size.

For logit-level distillation, we first construct label-conditioned margins:
\begin{equation}
m_i^S=(1-2y_i)s_i^S,
\qquad
m_i^T=(1-2y_i)s_i^T.
\end{equation}
Their temperature-scaled probabilities are defined as
  \begin{equation}
  \begin{aligned}
  \widetilde{p}_i^S &= \operatorname{clip}\!\left(\sigma(m_i^S/\tau),\,\delta,\,1-\delta\right), \\[2pt]
  \widetilde{p}_i^T &= \operatorname{clip}\!\left(\sigma(m_i^T/\tau),\,\delta,\,1-\delta\right).
  \end{aligned}
  \end{equation}
where $\tau$ is the distillation temperature, $\delta=10^{-6}$ ensures numerical stability, and $\sigma(\cdot)$ denotes the Sigmoid function. The logit-level distillation loss is then
\begin{equation}
\mathcal{L}_{\mathrm{logit}}
=
\frac{\tau^2}{B}
\sum_{i=1}^{B}
\operatorname{BCE}
\left(
\widetilde{p}_i^S,
\operatorname{sg}(\widetilde{p}_i^T)
\right),
\label{eq:logit_distillation}
\end{equation}
where $\operatorname{sg}(\cdot)$ denotes the stop-gradient operation, and the second argument of $\operatorname{BCE}(\cdot,\cdot)$ is treated as the target.

The sample-wise feature discrepancy is computed in the shared
projection space:
\begin{equation}
\ell_{\mathrm{feat}}^{(i)}
=
\frac{1}{d}
\left\|
\mathbf{u}_i^S-\mathbf{u}_i^T
\right\|_2^2,
\end{equation}
where $d$ denotes the feature dimensionality. The reliability-weighted
feature distillation loss is defined as
\begin{equation}
\mathcal{L}_{\mathrm{feat}}
=
\frac{
\sum_{i=1}^{B}
\operatorname{sg}(w_i)\,
\ell_{\mathrm{feat}}^{(i)}
}{
\sum_{i=1}^{B}w_i+\delta
}.
\label{eq:weighted_feature_loss}
\end{equation}
Here, $\operatorname{sg}(\cdot)$ denotes the stop-gradient operation. It is applied to the gate weights in the numerator, preventing the gate from trivially suppressing samples with large feature discrepancies to reduce the distillation loss. The normalization denominator remains differentiable and does not induce a downward collapse of the gate weights. Meanwhile, the mean-anchoring regularizer in Eq.~\eqref{eq:gate_regularization} constrains the average gate activation around $\rho$, preventing the gate from drifting toward either uniformly small or uniformly large values.

\paragraph{Progressive Optimization Objective.}
The overall training objective evolves with the training stage and is defined as
\begin{equation}
\begin{aligned}
\mathcal{L}_{\mathrm{total}}
={}&
\mathcal{L}_{\mathrm{task}}
+
\mathbb{I}(e>E_0)
\lambda_l
\mathcal{L}_{\mathrm{logit}}
\\
&+
\mathbb{I}(e>E_1)
\left[
\lambda_f
\mathcal{L}_{\mathrm{feat}}
+
\lambda_g
\mathcal{L}_{\mathrm{gate}}
\right],
\end{aligned}
\label{eq:total_objective}
\end{equation}
where $e$ denotes the current epoch, $E_0$ and $E_1$ are the stage-transition epochs, and $\mathbb{I}(\cdot)$ is the indicator function. The coefficients $\lambda_l$, $\lambda_f$, and $\lambda_g$ control the logit distillation, feature distillation, and gate regularization terms, respectively.

\section{Experiments and Results}

\subsection{Experimental Setup}

We evaluate BioKD on two widely used multimodal emotion recognition benchmarks, DEAP~\cite{koelstra2011deap} and AMIGOS~\cite{miranda2018amigos}. 
DEAP contains physiological signals and facial videos from 32 subjects, while AMIGOS provides multimodal recordings from 40 subjects; the short-video subset is used in our experiments.

Valence and arousal labels are binarized into high and low classes using a threshold of 5. We consider two data partition protocols. Under the \textbf{trial-wise} protocol, complete trials are assigned to mutually exclusive training, validation, and test sets with a ratio of 8:1:1. Under the \textbf{subject-wise} protocol, subjects are assigned to mutually exclusive splits using the same ratio, ensuring no subject overlap across the three sets.

To prevent temporal leakage, dataset partitioning is performed before window extraction. After the splits are determined, physiological signals and video streams are independently segmented within each split using a sliding window of 4 seconds with a stride of 2 seconds. Consequently, overlapping or adjacent windows from the same trial never appear in different splits.

Under both partition protocols, all experiments are repeated over 3 random seeds to assess robustness and result stability.
%Accuracy is reported as the primary evaluation metric.
We report accuracy and F1-score as mean$\pm$standard deviation over three random seeds.

To examine robustness under different levels of privileged information, we evaluate multiple teacher modality configurations.
On DEAP, teacher models range from EEG-only to full multimodal physiological settings.
On AMIGOS, teacher models are likewise expanded from EEG-only to full multimodal physiological settings.
All experiments are conducted for both valence and arousal tasks under the same protocol.

\begin{table}[t]
\centering
\caption{Architectural configuration of the physiological teacher and
video student. The teacher is used only during training.}
\label{tab:network_architecture}
\small
\setlength{\tabcolsep}{3.5pt}
\renewcommand{\arraystretch}{1.12}
\begin{tabular}{llcc}
\toprule
\textbf{Branch}
& \textbf{Component}
& \textbf{Output}
& \textbf{Status}
\\
\midrule

Teacher
& Modality encoders
& $M \times d_m$
& Frozen
\\

Teacher
& Temporal encoder
& $M \times d_t$
& Frozen
\\

Teacher
& MoE fusion
& 128
& Frozen
\\

Teacher
& Classifier
& 1
& Frozen
\\
\midrule

Student
& MobileNetV2
& $T \times d_v$
& Trainable
\\

Student
& GRU
& $d_s$
& Trainable
\\

Shared
& Trend projector
& 128
& Training only
\\

Student
& Classifier
& 1
& Trainable
\\

\bottomrule
\end{tabular}
\end{table}

\paragraph{Network Architectures}
We adopt an asymmetric teacher--student architecture, whose detailed configuration is summarized in Table~\ref{tab:network_architecture}. The physiological teacher takes synchronized physiological windows as input and employs modality-specific encoders, temporal modeling, and mixture-of-experts fusion to obtain an affective representation $\mathbf{z}_i^T$ and the corresponding prediction logits. The teacher is initialized with pre-trained weights and remains frozen throughout distillation. The video student receives only facial video frames without audio input, since speech may be absent during passive viewing and recorded audio may be dominated by stimulus or environmental sound. MobileNetV2~\cite{sandler2018mobilenetv2} extracts frame-level visual features, which are subsequently aggregated by a GRU to obtain the video representation $\mathbf{z}_i^S$. A shared projection head $\phi(\cdot)$ maps both teacher and student representations into a common 128-dimensional space for feature-level distillation. The projection head is used only during training and is discarded during inference. The student is trained from scratch and is the only network retained during inference.
Unless otherwise specified, MobileNetV2--GRU is used as the default student architecture; VideoMAE is evaluated separately in
Section~\ref{sec:backbone_generalization}.

\paragraph{Implementation Details}
All models are implemented in PyTorch and trained on an RTX A6000 GPU.
We use Adam with a learning rate of $3\times10^{-4}$ and a batch size of 16. The memory momentum is set to $m=0.92$, and the minimum gate weight is set to $\epsilon=0.01$. We adopt a three-stage progressive training schedule over 10 epochs:
(i) task-only learning in epochs 1--2,
(ii) logit-level distillation in epochs 3--5, and
(iii) full reliability-aware feature distillation in epochs 6--10.

\paragraph{Baselines}
To validate the effectiveness of the proposed BioKD framework, we reproduce several representative knowledge distillation methods as baselines under a cross-modal experimental setup, including KD~\cite{hinton2015distilling}, Fitnets~\cite{romero2014fitnets}, VID~\cite{ahn2019variational}, RKD~\cite{park2019relational}, EmotionKD~\cite{liu2023emotionkd}, C2KD~\cite{huo2024c2kd} and GRCGD~\cite{wu2025cross}, as shown in Table
\ref{tab:subjectwise_results} and Table ~\ref{tab:trialwise_results}.
In addition, a Student Only model trained without any teacher supervision is included as a baseline.
Unless otherwise specified, the teacher model used for distillation is trained on EEG signals only. The impact of different teacher modality configurations is studied separately in Table~\ref{tab:teacher_modality}.

For the calibration analysis, we additionally include Similarity-Preserving Knowledge Distillation (SP)~\cite{tung2019similarity} and Neuron Selectivity Transfer (NST)~\cite{huang2017like}.

\begin{table*}[t]
\centering
\caption{
Quantitative comparison under the trial-wise protocol on DEAP and AMIGOS.
Entries report Accuracy and F1-score as mean$\pm$standard
deviation (\%). Bold indicates the best result among video-based student models.
}
\label{tab:trialwise_results}
\renewcommand{\arraystretch}{0.95}
\resizebox{0.95\textwidth}{!}{%
\begin{tabular}{lcccccccc}
\toprule
\multirow{3}{*}{\textbf{Method}}
& \multicolumn{4}{c}{\textbf{DEAP}}
& \multicolumn{4}{c}{\textbf{AMIGOS}} \\
\cmidrule(lr){2-5} \cmidrule(lr){6-9}
& \multicolumn{2}{c}{\textbf{Accuracy}}
& \multicolumn{2}{c}{\textbf{F1-Score}}
& \multicolumn{2}{c}{\textbf{Accuracy}}
& \multicolumn{2}{c}{\textbf{F1-Score}} \\
\cmidrule(lr){2-3} \cmidrule(lr){4-5}
\cmidrule(lr){6-7} \cmidrule(lr){8-9}
& \textbf{Valence} & \textbf{Arousal}
& \textbf{Valence} & \textbf{Arousal}
& \textbf{Valence} & \textbf{Arousal}
& \textbf{Valence} & \textbf{Arousal} \\
\midrule

\multicolumn{9}{l}{\textit{Baselines}} \\

Physiological Teacher (EEG)
& \textcolor{gray}{61.48$\pm$4.51}
& \textcolor{gray}{64.49$\pm$5.57}
& \textcolor{gray}{62.84$\pm$6.21}
& \textcolor{gray}{64.87$\pm$8.06}
& \textcolor{gray}{54.33$\pm$6.34}
& \textcolor{gray}{65.66$\pm$7.18}
& \textcolor{gray}{59.86$\pm$6.71}
& \textcolor{gray}{64.51$\pm$7.65} \\

Video Student (No KD)
& 70.87$\pm$3.99
& 65.14$\pm$5.03
& 72.69$\pm$5.24
& 72.14$\pm$6.94
& 55.42$\pm$5.71
& 60.63$\pm$6.72
& 58.28$\pm$6.03
& 63.55$\pm$7.15 \\

\midrule
\multicolumn{9}{l}{\textit{Distillation Methods}} \\

KD~\cite{hinton2015distilling}
& 68.38$\pm$5.31
& 53.10$\pm$7.11
& 69.21$\pm$5.67
& 56.82$\pm$7.68
& 54.07$\pm$6.02
& 61.08$\pm$7.34
& 56.11$\pm$6.28
& 62.40$\pm$7.82 \\

FitNets~\cite{romero2014fitnets}
& 72.98$\pm$6.13
& 51.96$\pm$8.27
& 74.18$\pm$6.48
& 57.91$\pm$8.81
& 55.65$\pm$6.44
& 59.33$\pm$8.54
& 58.76$\pm$6.72
& 60.07$\pm$9.06 \\

VID~\cite{ahn2019variational}
& 71.98$\pm$5.62
& 55.80$\pm$7.62
& 71.62$\pm$5.97
& 58.06$\pm$8.18
& 57.91$\pm$5.89
& 60.32$\pm$7.85
& 59.24$\pm$6.14
& 64.01$\pm$8.34 \\

RKD~\cite{park2019relational}
& 65.44$\pm$6.57
& 64.69$\pm$9.12
& 67.86$\pm$6.91
& 66.02$\pm$9.68
& 53.11$\pm$6.83
& 53.03$\pm$9.34
& 56.90$\pm$6.96
& 56.97$\pm$9.81 \\

EmotionKD~\cite{liu2023emotionkd}
& 71.55$\pm$4.78
& 55.60$\pm$6.52
& 72.36$\pm$5.12
& 60.36$\pm$7.01
& 57.48$\pm$5.02
& 61.50$\pm$6.73
& 60.04$\pm$5.38
& 63.13$\pm$7.26 \\

C2KD~\cite{huo2024c2kd}
& 69.55$\pm$4.61
& 54.11$\pm$6.38
& 71.84$\pm$4.94
& 55.42$\pm$6.84
& 51.90$\pm$5.15
& 61.11$\pm$6.57
& 54.62$\pm$5.44
& 64.86$\pm$7.03 \\

GRCGD~\cite{wu2025cross}
& 72.53$\pm$5.08
& 53.45$\pm$6.91
& 73.46$\pm$5.46
& 56.79$\pm$7.43
& 57.46$\pm$5.34
& 61.31$\pm$7.12
& 61.02$\pm$5.71
& 62.08$\pm$7.58 \\

\midrule

\textbf{BioKD (EEG $\rightarrow$ Video) (Ours)}
& \textbf{74.00$\pm$3.16}
& \textbf{68.01$\pm$5.18}
& \textbf{76.74$\pm$3.32}
& \textbf{74.60$\pm$4.86}
& \textbf{59.32$\pm$3.58}
& \textbf{64.63$\pm$4.51}
& \textbf{63.85$\pm$3.96}
& \textbf{69.10$\pm$5.03} \\

\bottomrule
\end{tabular}%
}
\end{table*}

\begin{table*}[t]
\centering
\caption{
Quantitative comparison under the subject-wise protocol on DEAP and AMIGOS.
Entries report Accuracy and F1-Score as mean$\pm$standard deviation (\%).
Bold indicates the best result among video-based student models.
}
\label{tab:subjectwise_results}
\renewcommand{\arraystretch}{0.95}
\resizebox{0.95\textwidth}{!}{%
\begin{tabular}{lcccccccc}
\toprule
\multirow{3}{*}{\textbf{Method}}
& \multicolumn{4}{c}{\textbf{DEAP}}
& \multicolumn{4}{c}{\textbf{AMIGOS}} \\
\cmidrule(lr){2-5} \cmidrule(lr){6-9}
& \multicolumn{2}{c}{\textbf{Accuracy}}
& \multicolumn{2}{c}{\textbf{F1-Score}}
& \multicolumn{2}{c}{\textbf{Accuracy}}
& \multicolumn{2}{c}{\textbf{F1-Score}} \\
\cmidrule(lr){2-3} \cmidrule(lr){4-5}
\cmidrule(lr){6-7} \cmidrule(lr){8-9}
& \textbf{Valence} & \textbf{Arousal}
& \textbf{Valence} & \textbf{Arousal}
& \textbf{Valence} & \textbf{Arousal}
& \textbf{Valence} & \textbf{Arousal} \\
\midrule

\multicolumn{9}{l}{\textit{Baselines}} \\

Physiological Teacher (EEG)
& \textcolor{gray}{51.94$\pm$8.21}
& \textcolor{gray}{59.06$\pm$10.42}
& \textcolor{gray}{54.54$\pm$8.77}
& \textcolor{gray}{65.01$\pm$11.36}
& \textcolor{gray}{55.21$\pm$6.18}
& \textcolor{gray}{66.98$\pm$8.85}
& \textcolor{gray}{60.30$\pm$6.92}
& \textcolor{gray}{71.88$\pm$9.47} \\

Video Student (No KD)
& 50.74$\pm$7.11
& 56.43$\pm$9.24
& 53.01$\pm$7.84
& 62.58$\pm$10.07
& 51.20$\pm$6.52
& 55.71$\pm$8.31
& 55.12$\pm$7.08
& 59.36$\pm$8.74 \\

\midrule
\multicolumn{9}{l}{\textit{Distillation Methods}} \\

KD~\cite{hinton2015distilling}
& 53.34$\pm$8.03
& 56.28$\pm$9.68
& 57.16$\pm$8.41
& 61.94$\pm$10.22
& 54.86$\pm$6.76
& 55.36$\pm$7.69
& 58.31$\pm$7.25
& 59.02$\pm$8.69 \\

FitNets~\cite{romero2014fitnets}
& 51.24$\pm$9.27
& 52.89$\pm$11.48
& 54.88$\pm$9.81
& 58.47$\pm$12.06
& 44.07$\pm$8.36
& 58.73$\pm$9.72
& 49.35$\pm$8.92
& 61.24$\pm$10.41 \\

VID~\cite{ahn2019variational}
& 54.09$\pm$8.62
& 51.08$\pm$10.73
& 56.72$\pm$9.14
& 57.36$\pm$11.39
& 56.70$\pm$7.42
& 58.19$\pm$9.18
& 60.18$\pm$8.03
& 62.73$\pm$9.86 \\

RKD~\cite{park2019relational}
& 50.10$\pm$9.64
& 57.13$\pm$12.21
& 53.42$\pm$9.93
& 63.08$\pm$12.68
& 50.33$\pm$8.71
& 57.48$\pm$10.34
& 54.69$\pm$8.88
& 61.81$\pm$10.82 \\

EmotionKD~\cite{liu2023emotionkd}
& 57.74$\pm$7.28
& 61.07$\pm$9.85
& 61.29$\pm$7.91
& 66.84$\pm$10.36
& 47.20$\pm$7.13
& 59.59$\pm$8.64
& 51.68$\pm$7.67
& 62.16$\pm$9.21 \\

C2KD~\cite{huo2024c2kd}
& 54.27$\pm$8.11
& 59.30$\pm$10.06
& 58.65$\pm$8.55
& 62.11$\pm$10.69
& 51.42$\pm$6.89
& 58.54$\pm$8.97
& 55.06$\pm$6.88
& 62.82$\pm$9.54 \\

GRCGD~\cite{wu2025cross}
& 54.07$\pm$7.83
& 60.23$\pm$10.51
& 56.31$\pm$8.28
& 65.42$\pm$11.08
& 50.59$\pm$7.54
& 57.26$\pm$8.76
& 54.92$\pm$7.96
& 60.14$\pm$9.33 \\

\midrule

\textbf{BioKD (EEG $\rightarrow$ Video) (Ours)}
& \textbf{59.13$\pm$7.42}
& \textbf{65.29$\pm$9.16}
& \textbf{70.69$\pm$8.08}
& \textbf{76.91$\pm$9.73}
& \textbf{57.05$\pm$6.63}
& \textbf{60.95$\pm$7.82}
& \textbf{60.65$\pm$7.14}
& \textbf{63.55$\pm$8.31} \\

\bottomrule
\end{tabular}%
}
\end{table*}

\subsection{Comparison Study}

Tables~\ref{tab:trialwise_results} and~\ref{tab:subjectwise_results} report the experimental results on DEAP and AMIGOS under the trial-wise and subject-wise protocols, respectively. Both Accuracy and F1-score are reported as mean$\pm$standard deviation, and the EEG-based physiological teacher is used for all distillation methods. Overall, BioKD achieves the highest mean classification accuracy among the evaluated video-based student models across all datasets, emotion dimensions, and evaluation protocols. Notably, under the more challenging subject-wise protocol, BioKD continues to provide clear mean improvements, indicating improved robustness under subject-independent evaluation.

Compared with the best competing methods under each setting, BioKD maintains consistent gains in mean accuracy. For example, on DEAP trial-wise arousal, BioKD improves the mean accuracy from RKD's 64.69\% to \textbf{68.01\%}. On AMIGOS trial-wise arousal, BioKD achieves \textbf{64.63\%}, compared with 61.50\% for EmotionKD. Under the more challenging DEAP subject-wise arousal setting, BioKD reaches \textbf{65.29\%}, compared with 61.07\% for EmotionKD. In addition, BioKD consistently improves over the Video Student (No KD) baseline under both evaluation protocols, indicating that physiological teacher supervision can enhance video-based emotion recognition.

Examining the behavior of different distillation methods reveals that traditional knowledge distillation techniques exhibit varying levels of effectiveness in the considered \textbf{physiology-to-video distillation} setting. Standard KD and some feature-alignment-based methods even lead to performance degradation in certain cases, suggesting that when the teacher model is built upon noisy physiological signals with substantial inter-subject variability, applying uniform distillation constraints may propagate unreliable supervision to the student model. In contrast, methods such as FitNets, VID, EmotionKD, C2KD, and GRCGD achieve relatively better performance in some tasks, indicating that feature transfer, sample selection, or relational modeling can partially alleviate cross-modal discrepancies. However, these approaches generally do not explicitly model the \textbf{sample-wise reliability} of the teacher supervision, causing their improvements to depend heavily on specific datasets, tasks, or evaluation protocols. In comparison, BioKD introduces a reliability-aware gating mechanism to adaptively regulate the distillation strength at the sample level, enabling more stable and effective cross-modal knowledge transfer across datasets and evaluation settings.

\subsection{Ablation Study}

Table~\ref{tab:ablation_arousal} presents the ablation results for the DEAP arousal task under the subject-wise protocol, where all variants are distilled from the same EEG-based teacher for a fair comparison. The student-only baseline already provides a reasonable starting point, while na\"ive logits distillation leads to a performance drop, confirming that directly transferring supervision from noisy physiological teachers can be harmful. 

Compared with logits KD, na\"ive feature distillation improves performance, suggesting that feature-level guidance provides a more informative and transferable supervision signal in the considered cross-modal setting. Introducing the reliability-aware gating mechanism further improves the results, indicating the importance of selectively filtering teacher supervision according to sample-wise reliability. Removing the memory component causes a noticeable degradation, showing that temporal reliability accumulation helps stabilize the gating behavior across samples. Likewise, removing the progressive training strategy leads to a clear performance drop, demonstrating that gradually introducing teacher supervision is important for stable optimization.

The component-wise gate ablations further show that different reliability cues contribute differently to the final performance. In particular, removing the agreement reward or the confidence term causes larger degradation, suggesting that these two factors play a critical role in identifying trustworthy supervision. Removing the disagreement penalty also reduces performance, although to a relatively smaller extent. Overall, the full BioKD model achieves the best result, validating the effectiveness and complementarity of all components in the proposed framework.

Beyond the component-wise ablations, we further examine two implementation choices in BioKD: the distance function used to measure prediction disagreement and the objective to which the reliability gate is applied. The results are summarized in Table~\ref{tab:design_ablation}. As shown in Table~\ref{tab:design_ablation}, the default L1 disagreement measure outperforms the L2 and cosine alternatives, suggesting that absolute probability differences provide a more stable decision-level conflict signal in the considered cross-modal setting. Applying the gate only to feature-level distillation achieves the best performance, whereas applying it to logit-level distillation or both objectives leads to lower results. This finding indicates that deep cross-modal feature alignment is particularly sensitive to unreliable physiological supervision, while logit-level alignment is comparatively less dependent on sample-wise gating.

\newcolumntype{Y}{>{\footnotesize\raggedright\arraybackslash}X}
\newcolumntype{Z}{>{\footnotesize\centering\arraybackslash}p{0.9cm}}

\begin{table}[t]
    \centering
    \scriptsize
    \caption{\textbf{Ablation Study of BioKD Components on DEAP (Arousal) under the Subject-wise Protocol.}
    All variants are distilled from the same EEG-based physiological teacher.}
    \label{tab:ablation_arousal}

    \setlength{\tabcolsep}{3pt}
    \renewcommand{\arraystretch}{1.08}

    \begin{tabularx}{\columnwidth}{
        Y
        |>{\centering\arraybackslash}p{0.38cm}
         >{\centering\arraybackslash}p{0.50cm}
         >{\centering\arraybackslash}p{0.62cm}
         >{\centering\arraybackslash}p{0.40cm}
         >{\centering\arraybackslash}p{0.40cm}
        |Z
    }
        \toprule
        {\footnotesize Variant} & Stage & $\mathcal{L}_{KD}$ & $\mathcal{L}_{Feat}$ & Gate & Mem & {\footnotesize Acc} \\
        \midrule
        Student Baseline         &  &  &  &  &  & 56.43 \\
        Logits KD                & \cmark & \cmark &  &  &  & 50.28 \\
        Na\"ive Feature KD       & \cmark & \cmark & \cmark &  &  & 52.91 \\
        w/o Memory               & \cmark & \cmark & \cmark & \cmark &  & 63.43 \\
        w/o Stage                &  & \cmark & \cmark & \cmark & \cmark & 62.36 \\
        w/o Agreement Reward     & \cmark & \cmark & \cmark & \cmark & \cmark & 56.88 \\
        w/o Disagreement Penalty & \cmark & \cmark & \cmark & \cmark & \cmark & 62.59 \\
        w/o Confidence Term      & \cmark & \cmark & \cmark & \cmark & \cmark & 57.55 \\
        \midrule
        \textbf{BioKD (Full)}    & \cmark & \cmark & \cmark & \cmark & \cmark & \textbf{65.29} \\
        \bottomrule
    \end{tabularx}
\end{table}

\begin{table}[t]
\centering
\caption{Ablation of disagreement metrics and gate placement on DEAP (Arousal) under the subject-wise protocol.}
\label{tab:design_ablation}
\small
\setlength{\tabcolsep}{5pt}
\begin{tabular}{llcc}
\toprule
\textbf{Design} & \textbf{Variant} & \textbf{Acc. (\%)} & \textbf{F1-Score (\%)} \\
\midrule
\multirow{3}{*}{Disagreement}
& L1 (default) & \textbf{65.29} & \textbf{76.91} \\
& L2 & 52.71 & 63.54 \\
& Cosine & 61.90 & 72.47 \\
\midrule
\multirow{4}{*}{Gate placement}
& No gate & 54.16 & 66.61 \\
& Logit only & 62.99 & 73.49 \\
& Feature only & \textbf{65.29} & \textbf{76.91} \\
& Both & 60.85 & 71.80 \\
\bottomrule
\end{tabular}
\end{table}

\begin{figure}
    \centering
    \includegraphics[width=0.9\linewidth]{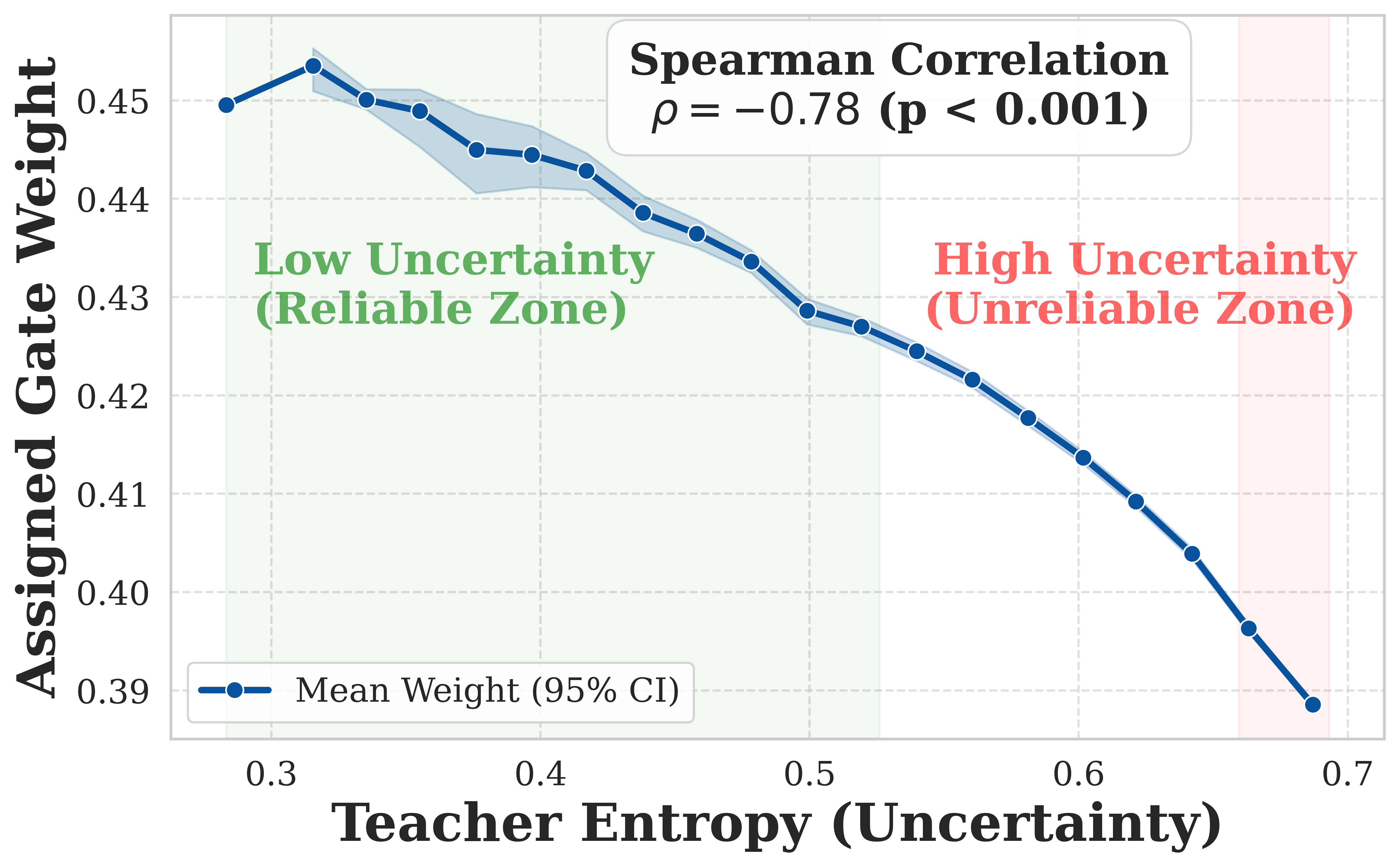}
    \caption{Relationship between teacher prediction entropy and the learned gate weight. The curve and shaded region denote the mean and 95\% confidence interval, respectively.}
    \label{fig:gating_analysis}
\end{figure}

\subsection{Generalization across Student Backbones}
\label{sec:backbone_generalization}

To examine whether the effectiveness of BioKD depends on the relatively lightweight MobileNetV2--GRU student, we further instantiate the video student using VideoMAE. For each backbone, we compare student-only training, standard knowledge distillation, and BioKD under the same data splits and evaluation protocol. As shown in Table~\ref{tab:backbone_generalization}, BioKD consistently improves both student architectures, indicating that its benefit arises from reliability-aware knowledge transfer rather than a particular video backbone.

\begin{table}[t]
\centering
\caption{Generalization of BioKD across different video student backbones on DEAP (Arousal) under the subject-wise protocol.}
\label{tab:backbone_generalization}
\small
\setlength{\tabcolsep}{4pt}
\begin{tabular}{llcc}
\toprule
\textbf{Backbone}
& \textbf{Training}
& \textbf{Acc}
& \textbf{F1-Score}
\\
\midrule
\multirow{3}{*}{MobileNetV2--GRU}
& Video Student (No KD) & 56.43 & 62.58 \\
& Standard KD & 56.28 & 61.94 \\
& BioKD & \textbf{65.29} & \textbf{76.91} \\
\midrule
\multirow{3}{*}{VideoMAE}
& Video Student (No KD) & 59.63 & 61.35 \\
& Standard KD  & 63.07 & 64.66 \\
& BioKD        & \textbf{66.78} & \textbf{70.73} \\
\bottomrule
\end{tabular}
\end{table}

\subsection{In-depth Analysis of Distillation Mechanism}
\label{sec:mechanism_analysis}

\subsubsection{Gate Reliability across Correctness Quadrants}
\label{sec:gate_reliability}

We evaluate whether the learned gate reflects teacher correctness and potential per-sample transfer utility rather than merely teacher--student agreement. Since the consistency memory is defined for training samples, this diagnostic analysis is conducted exclusively on the training split. For each sample, we record the gate weights produced during Stage~2 and average them across its training visits. The ground-truth labels are used only for post-hoc analysis and do not directly supervise the gate.

We then partition the samples into four correctness quadrants according to the physiological teacher prediction, the independently trained Video Student (No KD) prediction, and the ground-truth label. This partition distinguishes teacher-helpful cases, where the teacher is correct and the video student is wrong, from teacher-harmful cases, where the teacher is wrong and the video student is correct. Validation and test samples are not used to construct the memory statistics or the correctness quadrants in this analysis.

\begin{table}[t]
\centering
\caption{Gate weights across teacher--student correctness quadrants. $T$, $S$, and $Y$ denote the physiological teacher prediction, Video Student (No KD) prediction, and ground-truth label, respectively.}
\label{tab:teacher_student_quadrants}
\small
\setlength{\tabcolsep}{4pt}
\begin{tabular}{lccc}
\toprule
\textbf{Subset} & \textbf{Definition} & \textbf{Samples} & \textbf{Mean gate} \\
\midrule
Both correct & $T=Y,\ S=Y$ & 8,348 & 0.4445 \\
Teacher helpful & $T=Y,\ S\neq Y$ & 3,761 & 0.4274 \\
Teacher harmful & $T\neq Y,\ S=Y$ & 3,574 & 0.3911 \\
Both wrong & $T\neq Y,\ S\neq Y$ & 5,110 & 0.3984 \\
\bottomrule
\end{tabular}
\end{table}

As shown in Table~\ref{tab:teacher_student_quadrants}, both-correct samples receive the highest average gate weight, while teacher-harmful samples receive the lowest. Teacher-helpful samples also receive relatively high weights, indicating that BioKD preserves physiological supervision when it has the potential to correct the video student. Aggregating the four quadrants, teacher-correct samples receive higher average gate weights than teacher-wrong samples (0.4392 vs. 0.3954), and the gate distinguishes teacher correctness with an AUROC of 0.7021. These results show that the learned gate captures supervision utility beyond teacher--student agreement alone.

\subsubsection{Correction of Overconfident Teacher Errors}
\label{sec:overconfident_correction}

\begin{table}[t]
\centering
\caption{\textbf{Correction of Overconfident Wrong Teacher Predictions.}
Results are reported on DEAP (Arousal) under the subject-wise protocol using an EEG-based teacher.}
\label{tab:overconfident_correction}
\begin{tabular}{lcc}
\toprule
Method & Corrected / Total & Correction Rate (\%) \\
\midrule
Video Student (No KD) & 19 / 76 & 25.00 \\
KD & 14 / 76 & 18.42 \\
BioKD & \textbf{30 / 76} & \textbf{39.47} \\
\bottomrule
\end{tabular}
\end{table}

To further examine whether BioKD can mitigate the overconfident errors highlighted in Fig.~\ref{fig:reliability_gap}, we collect sample-level predictions from the teacher, Video Student (No KD), standard KD, and BioKD on the same test set. Specifically, we define overconfident wrong teacher samples as those on which the teacher prediction is incorrect and the prediction confidence, computed as the maximum posterior probability, exceeds 0.85. For this subset, we evaluate whether each student model can correct the teacher’s error, i.e., whether the student prediction matches the ground-truth label.
As shown in Table~\ref{tab:overconfident_correction}, among the 76 overconfident wrong teacher samples, BioKD successfully corrects 30 cases (39.47\%), outperforming both Video Student (No KD) (25.00\%) and standard KD (18.42\%). This result suggests that the proposed reliability-aware gating mechanism can effectively suppress misleading teacher supervision and reduce the tendency of the student to inherit overconfident teacher errors.

%\subsubsection{Analysis of Reliability-Aware Gating}
\subsubsection{Beyond Teacher Confidence}
\label{sec:gating_analysis}

We first examine how the learned gate relates to teacher prediction uncertainty. This analysis characterizes the behavior of the gate but does not by itself establish supervision reliability, since a confident teacher prediction may still be incorrect. Fig.~\ref{fig:gating_analysis} shows the relationship between teacher prediction entropy and the average gate weight.

As shown in the figure, the gate weight exhibits a decreasing trend as teacher entropy increases, and this relationship is further supported by a negative Spearman correlation of $\rho=-0.78$ with $p<0.001$. This result indicates that uncertain teacher predictions tend to receive weaker supervision weights. Nevertheless, the correctness-based analyses in Section~\ref{sec:gate_reliability} show that the gate behavior cannot be explained by confidence alone.

To further examine whether the proposed gating mechanism can be reduced to a simple entropy-based weighting strategy, we implement an \textit{entropy-only} baseline, where the distillation weights are computed solely from teacher prediction entropy. Under the subject-wise protocol on DEAP, the entropy-only baseline achieves 55.36\% and 59.48\% accuracy on the valence and arousal tasks, respectively. While it performs better than uniformly weighted standard KD, it still underperforms BioKD, which reaches 59.13\% and 65.29\% under the same setting. This indicates that teacher confidence alone is insufficient to fully characterize supervision reliability.

\begin{table}[t]
\centering
\caption{Sensitivity to the progressive distillation schedule on DEAP (Arousal) under the subject-wise protocol. Stage allocation reports the numbers of epochs assigned to Stages 0, 1, and 2, respectively.}
\label{tab:schedule_sensitivity}
\small
\setlength{\tabcolsep}{4pt}
\begin{tabular}{lcccccc}
\toprule
\textbf{Setting} & \(\boldsymbol{E_0}\) & \(\boldsymbol{E_1}\) & \textbf{Stage Allocation} & \textbf{Acc. (\%)} & \textbf{F1 (\%)} \\
\midrule
Early & 1 & 4 & 1/3/6 & 64.33 & 75.16 \\
Default & 2 & 5 & 2/3/5 & \textbf{65.29} & \textbf{76.91} \\
Late & 3 & 6 & 3/3/4 & 64.04 & 76.03 \\
\bottomrule
\end{tabular}
\end{table}

\subsubsection{Prediction Calibration}
\label{sec:calibration}

Beyond classification accuracy, we evaluate whether reliability-aware distillation improves the calibration of the video student. For each sample, prediction confidence is defined as the maximum posterior probability, i.e., $\max(p_i^S,1-p_i^S)$. Expected calibration error (ECE) is computed using 15 equally spaced confidence bins:
\begin{equation}
\operatorname{ECE}
=
\sum_{k=1}^{15}
\frac{|\mathcal{B}_k|}{N}
\left|
\operatorname{acc}(\mathcal{B}_k)
-
\operatorname{conf}(\mathcal{B}_k)
\right|,
\end{equation}
where $\mathcal{B}_k$ denotes the set of predictions assigned to the $k$-th confidence bin, and $N$ is the total number of evaluated samples. Lower ECE indicates better agreement between predictive confidence and empirical correctness. In addition to the main baselines, we include Similarity-Preserving Knowledge Distillation (SP)~\cite{tung2019similarity} and Neuron Selectivity Transfer (NST)~\cite{huang2017like} in this calibration analysis.

\begin{table}[t]
\centering
\caption{Expected calibration error on DEAP Arousal under the
subject-wise protocol. Lower values indicate better calibration.}
\label{tab:calibration}
\small
\begin{tabular}{lc}
\toprule
\textbf{Method} & \textbf{ECE} $\downarrow$ \\
\midrule
Video Student (No KD) & 0.234 \\
SP~\cite{tung2019similarity} & 0.215 \\
FitNets~\cite{romero2014fitnets} & 0.190 \\
VID~\cite{ahn2019variational} & 0.208 \\
NST~\cite{huang2017like} & 0.242 \\
BioKD & \textbf{0.089} \\
\bottomrule
\end{tabular}
\end{table}

As shown in Table~\ref{tab:calibration}, BioKD achieves the lowest ECE of 0.089, compared with 0.234 for the Video Student (No KD) and 0.190 for the best-performing competing distillation baseline, FitNets. This corresponds to ECE reductions of 62.0\% and 53.2\%, respectively, indicating that the confidence estimates produced by BioKD are better aligned with empirical correctness. This result suggests that reliability-aware distillation improves not only classification performance but also the calibration of the deployed video student.

\subsubsection{Qualitative Gate Analysis}
\label{sec:qualitative_gate}

\begin{figure}[t]
\centering
\includegraphics[width=\columnwidth]{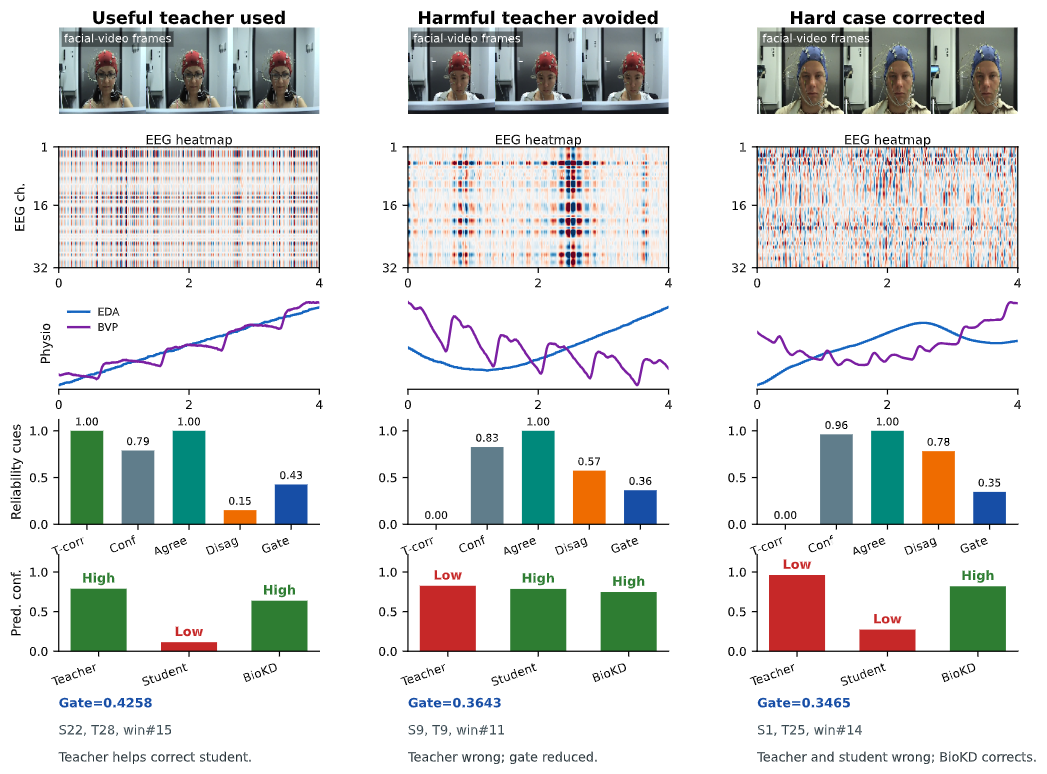}
\caption{Representative gate behaviors for helpful, harmful, and overconfident wrong-teacher cases.}
\label{fig:qualitative_gate_cases}
\end{figure}

To complement the aggregate analyses, we examine representative sample-level cases in which teacher supervision is useful, harmful, or overconfidently incorrect. At the group level, useful-teacher-used cases receive higher average gate weights than harmful-teacher-avoided cases (0.4265 vs. 0.3710). Moreover, overconfident wrong-teacher cases exhibit a high average teacher confidence of 0.7759 but receive a substantially lower average gate weight of 0.3950. Fig.~\ref{fig:qualitative_gate_cases} presents representative cases from these groups.

As illustrated in Fig.~\ref{fig:qualitative_gate_cases}, BioKD assigns relatively high gate weights when physiological teacher supervision provides useful corrective information and suppresses knowledge transfer when the teacher conflicts with an already correct video student. The overconfident wrong-teacher examples further demonstrate that the gate can remain conservative even when the teacher produces a strong but incorrect prediction.

%\begin{figure}
%    \centering
%    \includegraphics[width=0.9\linewidth]{figures/sensitivity_lambda_f_acc_DEAP_EEG_arousal.png}
%    \caption{Sensitivity to $\lambda_f$.}
    %\Description{
    %Sensitivity analysis of the feature distillation weight parameter $\lambda_f$.
    %Model performance varies with different values of $\lambda_f$, showing the impact of feature-level distillation strength.
    %}
%    \label{fig:sensitivity_lambda_f}
%\end{figure}

\begin{figure}[t]
\centering
\includegraphics[width=\columnwidth]{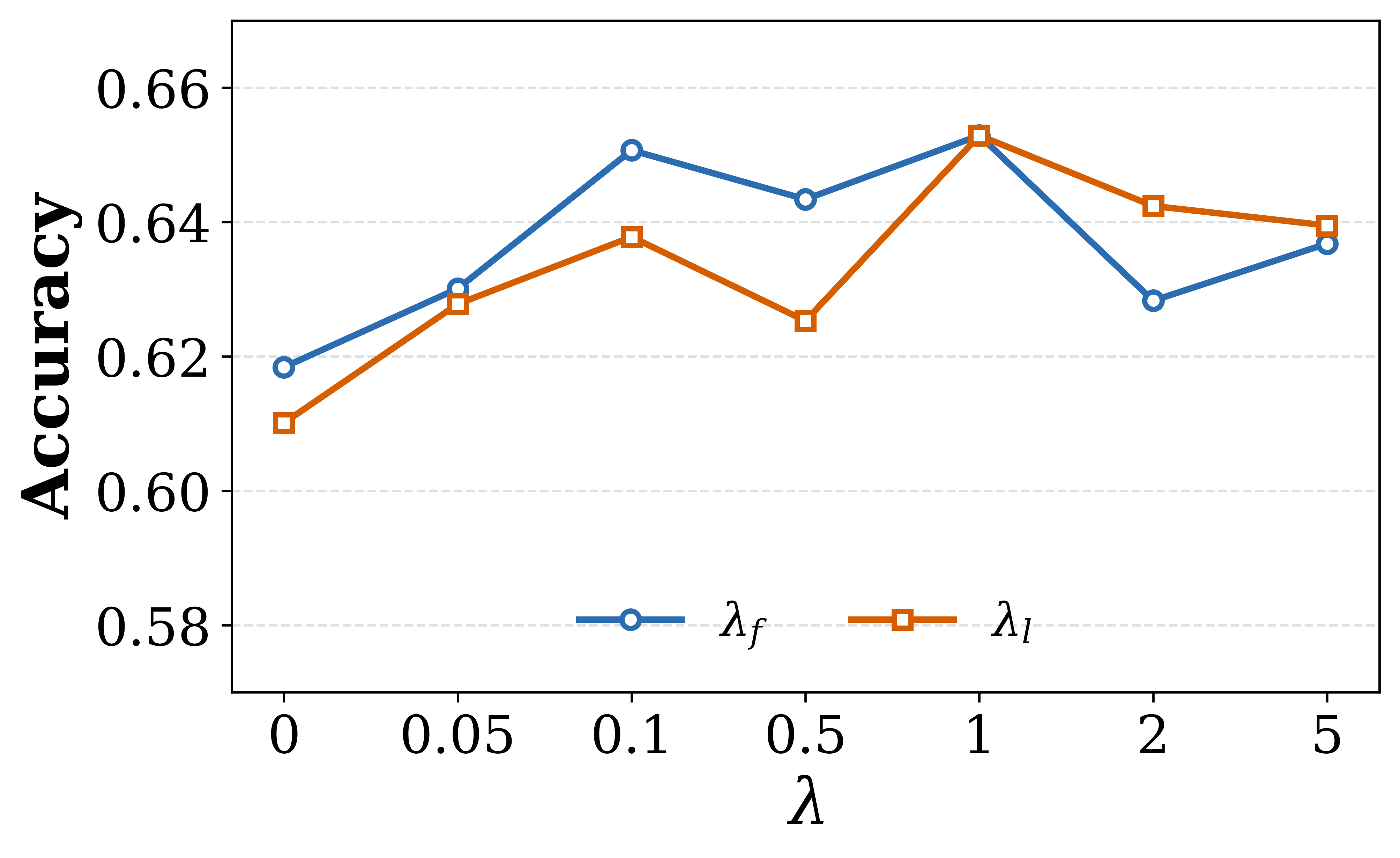}
\caption{Sensitivity of BioKD to the feature-level coefficient $\lambda_f$ and logit-level coefficient $\lambda_l$ on DEAP (Arousal) under the subject-wise protocol. When varying one coefficient, the other is fixed at its default value of $1.0$.}
\label{fig:distillation_sensitivity}
\end{figure}

\subsubsection{Hyperparameter Sensitivity Analysis}

%We evaluate the sensitivity of BioKD to the feature distillation weight $\lambda_f$ on the subject-wise DEAP arousal task, with results shown in Fig.~\ref{fig:sensitivity_lambda_f}. BioKD maintains stable performance across a broad range of $\lambda_f$ values, indicating low sensitivity to this hyperparameter. Moderate values of $\lambda_f$ achieve a balance between leveraging teacher feature information and mitigating noisy supervision, while excessively large values lead to performance degradation. Overall, these results demonstrate that BioKD does not require precise tuning of $\lambda_f$, supporting the robustness and practical usability of the proposed framework.

%We evaluate the sensitivity of BioKD to the feature-distillation coefficient $\lambda_f$, logit-distillation coefficient $\lambda_l$, and progressive schedule on DEAP (Arousal) under the subject-wise protocol. As shown in Fig.~\ref{fig:distillation_sensitivity}, BioKD maintains relatively stable performance across moderate values of both distillation coefficients, while excessively weak or strong supervision leads to lower performance. Table~\ref{tab:schedule_sensitivity} further compares different transition timings while keeping the duration of Stage 1 fixed at three epochs. The default schedule $(E_0,E_1)=(2,5)$ achieves the best overall result, whereas shifting the transitions earlier or later results in only moderate performance changes. These results indicate that BioKD benefits from a balanced progressive schedule without relying on a narrowly tuned transition point.

We evaluate the sensitivity of BioKD to the feature-distillation coefficient $\lambda_f$, the logit-distillation coefficient $\lambda_l$, and the progressive schedule on DEAP (Arousal) under the subject-wise protocol. As shown in Fig.~\ref{fig:distillation_sensitivity}, each coefficient is varied independently while the other is fixed at its default value of $1.0$. Although $\lambda_f$ and $\lambda_l$ exhibit different local sensitivity patterns because they regulate knowledge transfer at different representation levels, both achieve the highest accuracy at $1.0$. Weakening or removing either objective provides insufficient distillation supervision, whereas an excessively large coefficient may over-constrain the student to match modality-specific features or unreliable teacher predictions. Therefore, the two results obtained at a coefficient value of zero correspond to different training configurations. Table~\ref{tab:schedule_sensitivity} further compares different transition timings while keeping the duration of Stage 1 fixed at three epochs. The default schedule $(E_0,E_1)=(2,5)$ achieves the best overall result, whereas shifting the transitions earlier or later results in only moderate performance changes. These results indicate that BioKD benefits from balanced feature- and logit-level supervision together with a progressive schedule, without relying on a narrowly tuned transition point.

\subsection{Robustness to Teacher Modality Configurations}

The results in Table~\ref{tab:teacher_modality} show that BioKD provides stable and effective supervision under different teacher modality configurations. Even when only a uni-modal EEG teacher is used, the student still achieves competitive performance, indicating that BioKD does not rely on a specific or highly complex teacher design. Meanwhile, the best-performing configuration varies across tasks and datasets, suggesting that additional physiological modalities may offer complementary benefits, while a full-modality teacher is not necessarily the optimal choice.

\begin{table}[t]
    \centering
    \caption{\textbf{Impact of Teacher Modality Configurations.} 
%    We evaluate BioKD under different teacher settings, ranging from uni-modal (EEG) to multi-modal combinations.
    }
    \label{tab:teacher_modality}
    \resizebox{\linewidth}{!}{
        \begin{tabular}{l|c|c}
            \toprule
            %\multirow{2}{*}{\textbf{Teacher Configuration}} 
            \textbf{Teacher Configuration}
            & \textbf{Arousal} 
            & \textbf{Valence} \\
            %& \textbf{Acc} & \textbf{Acc} \\
            \midrule
            \multicolumn{3}{l}{\textit{DEAP Dataset}} \\
            1. EEG Only & 65.29 & \textbf{59.13} \\
            2. EEG + EMG + EOG & \textbf{69.78} & 55.83 \\
            3. EEG + BVP + EDA + TEMP + RESP & 65.23 & 56.36 \\
            4. All Modalities (Full) & 67.34 & 54.48 \\
            \midrule
            \multicolumn{3}{l}{\textit{AMIGOS Dataset}} \\
            1. EEG Only & \textbf{60.95} & 57.05 \\
            2. EEG + ECG & 58.75& \textbf{61.66} \\
            3. EEG + GSR & 57.79 & 56.39 \\
            4. EEG + GSR + ECG (Full) & 60.27 & 59.45 \\
            \bottomrule
        \end{tabular}
        }
\end{table}

\begin{table}[t]
\centering
\caption{Inference cost and deployment characteristics.}
\label{tab:efficiency}
\small
\renewcommand{\arraystretch}{1.12}
\begin{tabular*}{\columnwidth}{
@{\extracolsep{\fill}}lcccc@{}
}
\toprule
\textbf{Model}
& \textbf{Input}
& \shortstack{\textbf{Params}\\\textbf{(M)}}
& \shortstack{\textbf{FLOPs}\\\textbf{(G)}}
& \shortstack{\textbf{Latency}\\\textbf{(ms)}} \\
\midrule
Physiological Teacher
& Physiology
& 2.94
& 1.93
& 8.96 \\
Video Student
& Video
& 2.57
& 3.38
& 4.82 \\
\bottomrule
\end{tabular*}
\end{table}

\subsection{Inference Cost and Deployment Characteristics}
\label{sec:efficiency}

BioKD is designed to reduce the sensing and modality requirements of physiology-assisted emotion recognition rather than to compress the video backbone. The physiological teacher, reliability gate, consistency memory, and shared projection head are used only during training and are discarded afterward. Consequently, the deployed model consists solely of the video student and introduces no additional inference-time modules or computational cost relative to the same student trained without distillation.

As shown in Table~\ref{tab:efficiency}, the deployed video student contains 2.57M parameters and requires 3.38G FLOPs, with a measured forward-pass latency of 4.82 ms under the reported hardware setting. Although processing video requires more FLOPs than processing one-dimensional physiological signals (3.38G vs. 1.93G), BioKD does not require physiological sensing, teacher execution, or multimodal synchronization during deployment. Therefore, its practical benefit lies in enabling non-intrusive video-only inference without increasing the computational complexity of the underlying video student, rather than in compressing the student backbone.

\section{Discussion}

This work revisits a core challenge in affective computing. Video is easy to acquire and deploy, but as an external behavioral signal, it may fail to reflect internal emotional states~\cite{poria2017review,pantic2003toward,gross1993emotional}. On the other hand, physiological signals provide more direct affective information, but they are noisy, highly subject-dependent, and difficult to use in real-world deployment~\cite{critchley2013interaction,scarciglia2023physiological,dzedzickis2020human}. Therefore, the problem studied here is not simply how to distill knowledge across modalities, but how to exploit useful physiological information during training while avoiding the transfer of unreliable supervision to the student model.

The main finding of this work is that, in physiology-to-video distillation, the overall performance of the teacher does not necessarily reflect the reliability of its supervision at the sample level. Even when the physiological teacher performs well on average, it can still produce high-confidence but incorrect predictions on difficult or noisy samples. This helps explain why uniform distillation may become unstable and even lead to negative transfer. In contrast, BioKD improves distillation stability by explicitly modeling sample-wise supervision reliability and selectively regulating teacher guidance. This suggests that the key issue is not simply whether the teacher is strong on average, but whether its supervision should be trusted for each sample. This finding also clarifies the difference between BioKD and previous methods. Conventional KD and feature-based distillation methods mainly focus on knowledge transfer~\cite{hinton2015distilling,romero2014fitnets,park2019relational,gou2021knowledge}, while recent cross-modal methods often focus on alignment or sample selection~\cite{gupta2016cross,liu2023emotionkd,huo2024c2kd,wu2025cross}. However, our results suggest that reducing the modality gap does not necessarily guarantee reliable supervision when the teacher itself is built on noisy physiological signals. In this setting, the main challenge is not only how to transfer knowledge, but also when teacher supervision should be trusted.

More broadly, these results provide a different perspective on the relationship between external and internal signals in affective computing. They suggest that physiological signals may not be ideal as mandatory inference-time inputs, but they can still serve as training-only privileged information that complements video with internal affective cues. Although BioKD is still limited by teacher quality and proxy-based reliability modeling, the results indicate that explicitly modeling supervision reliability is necessary for stable physiology-to-video knowledge distillation.

\section{Conclusion}

This paper presented BioKD, a reliability-aware physiology-to-video knowledge distillation framework for video-based emotion recognition. By treating physiological signals as training-only privileged information, BioKD selectively transfers complementary affective knowledge while retaining facial video as the only inference-time input. Its sample-wise reliability gate and progressive distillation strategy jointly regulate unreliable physiological supervision and reduce negative transfer during cross-modal learning. Experiments on DEAP and AMIGOS demonstrated consistent improvements over representative distillation baselines under both trial-wise and subject-wise protocols. Further analyses confirmed the importance of distinguishing teacher confidence from supervision reliability. Future work will investigate more adaptive reliability estimation and broader generalization across datasets and video backbones.

\section*{Acknowledgments}
This work was supported by Guangdong Basic and Applied Basic Research Foundation under Grant No. 2025A1515110098.

\bibliographystyle{IEEEtran}
\bibliography{references}

@article{hinton2015distilling,
  title={Distilling the knowledge in a neural network},
  author={Hinton, Geoffrey and Vinyals, Oriol and Dean, Jeff},
  journal={arXiv preprint arXiv:1503.02531},
  year={2015}
}

@article{gou2021knowledge,
  title={Knowledge distillation: A survey},
  author={Gou, Jianping and Yu, Baosheng and Maybank, Stephen J and Tao, Dacheng},
  journal={International journal of computer vision},
  volume={129},
  number={6},
  pages={1789--1819},
  year={2021},
  publisher={Springer}
}

@inproceedings{park2019relational,
  title={Relational knowledge distillation},
  author={Park, Wonpyo and Kim, Dongju and Lu, Yan and Cho, Minsu},
  booktitle={Proceedings of the IEEE/CVF conference on computer vision and pattern recognition},
  pages={3967--3976},
  year={2019}
}

@article{zagoruyko2016paying,
  title={Paying more attention to attention: Improving the performance of convolutional neural networks via attention transfer},
  author={Zagoruyko, Sergey and Komodakis, Nikos},
  journal={arXiv preprint arXiv:1612.03928},
  year={2016}
}

@inproceedings{mirzadeh2020improved,
  title={Improved knowledge distillation via teacher assistant},
  author={Mirzadeh, Seyed Iman and Farajtabar, Mehrdad and Li, Ang and Levine, Nir and Matsukawa, Akihiro and Ghasemzadeh, Hassan},
  booktitle={Proceedings of the AAAI conference on artificial intelligence},
  volume={34},
  pages={5191--5198},
  year={2020}
}

@article{romero2014fitnets,
  title={Fitnets: Hints for thin deep nets. arXiv 2014},
  author={Romero, Adriana and Ballas, Nicolas and Kahou, Samira Ebrahimi and Chassang, Antoine and Gatta, Carlo and Bengio, Yoshua},
  journal={arXiv preprint arXiv:1412.6550},
  year={2014}
}

@inproceedings{gupta2016cross,
  title={Cross modal distillation for supervision transfer},
  author={Gupta, Saurabh and Hoffman, Judy and Malik, Jitendra},
  booktitle={Proceedings of the IEEE conference on computer vision and pattern recognition},
  pages={2827--2836},
  year={2016}
}

@inproceedings{liu2023emotionkd,
  title={Emotionkd: a cross-modal knowledge distillation framework for emotion recognition based on physiological signals},
  author={Liu, Yucheng and Jia, Ziyu and Wang, Haichao},
  booktitle={Proceedings of the 31st ACM international conference on multimedia},
  pages={6122--6131},
  year={2023}
}

@article{lawhern2018eegnet,
  title={EEGNet: a compact convolutional neural network for EEG-based brain--computer interfaces},
  author={Lawhern, Vernon J and Solon, Amelia J and Waytowich, Nicholas R and Gordon, Stephen M and Hung, Chou P and Lance, Brent J},
  journal={Journal of neural engineering},
  volume={15},
  number={5},
  pages={056013},
  year={2018},
  publisher={iOP Publishing}
}

@inproceedings{guo2017calibration,
  title={On calibration of modern neural networks},
  author={Guo, Chuan and Pleiss, Geoff and Sun, Yu and Weinberger, Kilian Q},
  booktitle={International conference on machine learning},
  pages={1321--1330},
  year={2017},
  organization={PMLR}
}

@article{yang2024uncertainty,
  title={Uncertainty-Aware Self-Knowledge Distillation},
  author={Yang, Yang and Wang, Chao and Gong, Lei and Wu, Min and Chen, Zhenghua and Gao, Yingxue and Wang, Teng and Zhou, Xuehai},
  journal={IEEE Transactions on Circuits and Systems for Video Technology},
  year={2024},
  publisher={IEEE}
}

@inproceedings{zhang2019your,
  title={Be your own teacher: Improve the performance of convolutional neural networks via self distillation},
  author={Zhang, Linfeng and Song, Jiebo and Gao, Anni and Chen, Jingwei and Bao, Chenglong and Ma, Kaisheng},
  booktitle={Proceedings of the IEEE/CVF international conference on computer vision},
  pages={3713--3722},
  year={2019}
}

@article{mansourian2025comprehensive,
  title={A Comprehensive Survey on Knowledge Distillation},
  author={Mansourian, Amir M and Ahmadi, Rozhan and Ghafouri, Masoud and Babaei, Amir Mohammad and Golezani, Elaheh Badali and Ghamchi, Zeynab Yasamani and Ramezanian, Vida and Taherian, Alireza and Dinashi, Kimia and Miri, Amirali and others},
  journal={arXiv preprint arXiv:2503.12067},
  year={2025}
}

@article{chen2024vision,
  title={Vision-language meets the skeleton: Progressively distillation with cross-modal knowledge for 3d action representation learning},
  author={Chen, Yang and He, Tian and Fu, Junfeng and Wang, Ling and Guo, Jingcai and Hu, Ting and Cheng, Hong},
  journal={IEEE Transactions on Multimedia},
  year={2024},
  publisher={IEEE}
}

@inproceedings{ahn2019variational,
  title={Variational information distillation for knowledge transfer},
  author={Ahn, Sungsoo and Hu, Shell Xu and Damianou, Andreas and Lawrence, Neil D and Dai, Zhenwen},
  booktitle={Proceedings of the IEEE/CVF conference on computer vision and pattern recognition},
  pages={9163--9171},
  year={2019}
}

@inproceedings{wu2025cross,
  title={A cross-modal densely guided knowledge distillation based on modality rebalancing strategy for enhanced unimodal emotion recognition},
  author={Wu, Shuang and Liang, Heng and Zhang, Yong and Chen, Yanlin and Jia, Ziyu},
  booktitle={Proceedings of the Thirty-Fourth International Joint Conference on Artificial Intelligence},
  pages={4236--4244},
  year={2025}
}

@article{poria2017review,
  title={A review of affective computing: From unimodal analysis to multimodal fusion},
  author={Poria, Soujanya and Cambria, Erik and Bajpai, Rajiv and Hussain, Amir},
  journal={Information fusion},
  volume={37},
  pages={98--125},
  year={2017},
  publisher={Elsevier}
}

@article{li2020deep,
  title={Deep facial expression recognition: A survey},
  author={Li, Shan and Deng, Weihong},
  journal={IEEE transactions on affective computing},
  volume={13},
  number={3},
  pages={1195--1215},
  year={2020},
  publisher={IEEE}
}

@article{gross1993emotional,
  title={Emotional suppression: physiology, self-report, and expressive behavior.},
  author={Gross, James J and Levenson, Robert W},
  journal={Journal of personality and social psychology},
  volume={64},
  number={6},
  pages={970},
  year={1993},
  publisher={American Psychological Association}
}

@article{pantic2003toward,
  title={Toward an affect-sensitive multimodal human-computer interaction},
  author={Pantic, Maja and Rothkrantz, Leon JM},
  journal={Proceedings of the IEEE},
  volume={91},
  number={9},
  pages={1370--1390},
  year={2003},
  publisher={IEEE}
}

@article{koelstra2011deap,
  title={Deap: A database for emotion analysis; using physiological signals},
  author={Koelstra, Sander and Muhl, Christian and Soleymani, Mohammad and Lee, Jong-Seok and Yazdani, Ashkan and Ebrahimi, Touradj and Pun, Thierry and Nijholt, Anton and Patras, Ioannis},
  journal={IEEE transactions on affective computing},
  volume={3},
  number={1},
  pages={18--31},
  year={2011},
  publisher={IEEE}
}

@incollection{critchley2013interaction,
  title={Interaction between cognition, emotion, and the autonomic nervous system},
  author={Critchley, Hugo D and Eccles, Jessica and Garfinkel, Sarah N},
  booktitle={Handbook of clinical neurology},
  volume={117},
  pages={59--77},
  year={2013},
  publisher={Elsevier}
}

@article{scarciglia2023physiological,
  title={Physiological noise: Definition, estimation, and characterization in complex biomedical signals},
  author={Scarciglia, Andrea and Catrambone, Vincenzo and Bonanno, Claudio and Valenza, Gaetano},
  journal={IEEE Transactions on Biomedical Engineering},
  volume={71},
  number={1},
  pages={45--55},
  year={2023},
  publisher={IEEE}
}

@article{lotte2018review,
  title={A review of classification algorithms for EEG-based brain--computer interfaces: a 10 year update},
  author={Lotte, Fabien and Bougrain, Laurent and Cichocki, Andrzej and Clerc, Maureen and Congedo, Marco and Rakotomamonjy, Alain and Yger, Florian},
  journal={Journal of neural engineering},
  volume={15},
  number={3},
  pages={031005},
  year={2018},
  publisher={iOP Publishing}
}

@article{vapnik2015learning,
  title={Learning using privileged information: Similarity control and knowledge transfer.},
  author={Vapnik, Vladimir and Izmailov, Rauf and others},
  journal={J. Mach. Learn. Res.},
  volume={16},
  number={1},
  pages={2023--2049},
  year={2015}
}

@inproceedings{aslam2023privileged,
  title={Privileged knowledge distillation for dimensional emotion recognition in the wild},
  author={Aslam, Muhammad Haseeb and Zeeshan, Muhammad Osama and Pedersoli, Marco and Koerich, Alessandro L and Bacon, Simon and Granger, Eric},
  booktitle={Proceedings of the IEEE/CVF conference on computer vision and pattern recognition},
  pages={3338--3347},
  year={2023}
}

@inproceedings{sandler2018mobilenetv2,
  title={Mobilenetv2: Inverted residuals and linear bottlenecks},
  author={Sandler, Mark and Howard, Andrew and Zhu, Menglong and Zhmoginov, Andrey and Chen, Liang-Chieh},
  booktitle={Proceedings of the IEEE conference on computer vision and pattern recognition},
  pages={4510--4520},
  year={2018}
}

@article{miranda2018amigos,
  title={Amigos: A dataset for affect, personality and mood research on individuals and groups},
  author={Miranda-Correa, Juan Abdon and Abadi, Mojtaba Khomami and Sebe, Nicu and Patras, Ioannis},
  journal={IEEE transactions on affective computing},
  volume={12},
  number={2},
  pages={479--493},
  year={2018},
  publisher={IEEE}
}

@inproceedings{huo2024c2kd,
  title={C2kd: Bridging the modality gap for cross-modal knowledge distillation},
  author={Huo, Fushuo and Xu, Wenchao and Guo, Jingcai and Wang, Haozhao and Guo, Song},
  booktitle={Proceedings of the IEEE/CVF Conference on Computer Vision and Pattern Recognition},
  pages={16006--16015},
  year={2024}
}

@article{zhu2025real,
  title={Real-time Cross-modal Cybersickness Prediction in Virtual Reality},
  author={Zhu, Yitong and Li, Tangyao and Wang, Yuyang},
  journal={arXiv preprint arXiv:2501.01212},
  year={2025}
}

@inproceedings{liu2019generalized,
  title={Generalized alignment for multimodal physiological signal learning},
  author={Liu, Yuchi and Yao, Yue and Wang, Zhengjie and Plested, Josephine and Gedeon, Tom},
  booktitle={2019 International Joint Conference on Neural Networks (IJCNN)},
  pages={1--10},
  year={2019},
  organization={IEEE}
}

@article{muller2019does,
  title={When does label smoothing help?},
  author={M{\"u}ller, Rafael and Kornblith, Simon and Hinton, Geoffrey E},
  journal={Advances in neural information processing systems},
  volume={32},
  year={2019}
}

@article{grathwohl2019your,
  title={Your classifier is secretly an energy based model and you should treat it like one},
  author={Grathwohl, Will and Wang, Kuan-Chieh and Jacobsen, J{\"o}rn-Henrik and Duvenaud, David and Norouzi, Mohammad and Swersky, Kevin},
  journal={arXiv preprint arXiv:1912.03263},
  year={2019}
}

@inproceedings{hoffman2016learning,
  title={Learning with side information through modality hallucination},
  author={Hoffman, Judy and Gupta, Saurabh and Darrell, Trevor},
  booktitle={Proceedings of the IEEE conference on computer vision and pattern recognition},
  pages={826--834},
  year={2016}
}

@article{ma2023transformer,
  title={A transformer-based model with self-distillation for multimodal emotion recognition in conversations},
  author={Ma, Hui and Wang, Jian and Lin, Hongfei and Zhang, Bo and Zhang, Yijia and Xu, Bo},
  journal={IEEE Transactions on Multimedia},
  volume={26},
  pages={776--788},
  year={2023},
  publisher={IEEE}
}

@article{hussain2025optimised,
  title={Optimised knowledge distillation for efficient social media emotion recognition using DistilBERT and ALBERT},
  author={Hussain, Muhammad and Chen, Caikou and Hussain, Muzammil and Anwar, Muhammad and Abaker, Mohammed and Abdelmaboud, Abdelzahir and Yamin, Iqra},
  journal={Scientific Reports},
  volume={15},
  number={1},
  pages={30104},
  year={2025},
  publisher={Nature Publishing Group UK London}
}

@article{zhu2025hierarchical,
  title={Hierarchical moe: Continuous multimodal emotion recognition with incomplete and asynchronous inputs},
  author={Zhu, Yitong and Han, Lei and Jiang, GuanXuan and Zhou, PengYuan and Wang, Yuyang},
  journal={arXiv preprint arXiv:2508.02133},
  year={2025}
}

@article{zhang2022visual,
  title={Visual-to-EEG cross-modal knowledge distillation for continuous emotion recognition},
  author={Zhang, Su and Tang, Chuangao and Guan, Cuntai},
  journal={Pattern Recognition},
  volume={130},
  pages={108833},
  year={2022},
  publisher={Elsevier}
}

@article{ahmed2019emotion,
  title={Emotion recognition from body movement},
  author={Ahmed, Ferdous and Bari, ASM Hossain and Gavrilova, Marina L},
  journal={IEEE Access},
  volume={8},
  pages={11761--11781},
  year={2019},
  publisher={IEEE}
}

@article{dzedzickis2020human,
  title={Human emotion recognition: Review of sensors and methods},
  author={Dzedzickis, Andrius and Kaklauskas, Art{\=u}ras and Bucinskas, Vytautas},
  journal={Sensors},
  volume={20},
  number={3},
  pages={592},
  year={2020},
  publisher={MDPI}
}

@article{ramaswamy2024multimodal,
  title={Multimodal emotion recognition: A comprehensive review, trends, and challenges},
  author={Ramaswamy, Manju Priya Arthanarisamy and Palaniswamy, Suja},
  journal={Wiley Interdisciplinary Reviews: Data Mining and Knowledge Discovery},
  volume={14},
  number={6},
  pages={e1563},
  year={2024},
  publisher={Wiley Online Library}
}

@article{wu2025comprehensive,
  title={A comprehensive review of multimodal emotion recognition: Techniques, challenges, and future directions},
  author={Wu, You and Mi, Qingwei and Gao, Tianhan},
  journal={Biomimetics},
  volume={10},
  number={7},
  pages={418},
  year={2025},
  publisher={MDPI}
}

@article{li2025multimodal,
  title={Multimodal physiological signals from wearable sensors for affective computing: A systematic review},
  author={Li, Fang and Zhang, Dan},
  journal={Intelligent Sports and Health},
  volume={1},
  number={4},
  pages={210--222},
  year={2025},
  publisher={Elsevier}
}

@article{ahmad2022survey,
  title={A survey on physiological signal-based emotion recognition},
  author={Ahmad, Zeeshan and Khan, Naimul},
  journal={Bioengineering},
  volume={9},
  number={11},
  pages={688},
  year={2022},
  publisher={MDPI}
}

@inproceedings{strizhkova2024mvp,
  title={Mvp: Multimodal emotion recognition based on video and physiological signals},
  author={Strizhkova, Valeriya and Kachmar, Hadi and Chaptoukaev, Hava and Kalandadze, Raphael and Kukhilava, Natia and Tsmindashvili, Tatia and Abo-Alzahab, Nibras and Zuluaga, Maria A and Balazia, Michal and Dantcheva, Antitza and others},
  booktitle={European Conference on Computer Vision},
  pages={101--116},
  year={2024},
  organization={Springer}
}

@inproceedings{lopez2024phemonet,
  title={PHemoNet: A multimodal network for physiological signals},
  author={Lopez, Eleonora and Uncini, Aurelio and Comminiello, Danilo},
  booktitle={2024 IEEE 8th Forum on Research and Technologies for Society and Industry Innovation (RTSI)},
  pages={260--264},
  year={2024},
  organization={IEEE}
}

@inproceedings{pan2023multimodal,
  title={Multimodal physiological signals fusion for online emotion recognition},
  author={Pan, Tongjie and Ye, Yalan and Cai, Hecheng and Huang, Shudong and Yang, Yang and Wang, Guoqing},
  booktitle={Proceedings of the 31st ACM international conference on multimedia},
  pages={5879--5888},
  year={2023}
}

@article{jia2025cross,
  title={Cross-modal knowledge distillation for enhanced unimodal emotion recognition},
  author={Jia, Ziyu and Liu, Yucheng and Wang, Haichao and Jiang, Tianzi},
  journal={IEEE Transactions on Affective Computing},
  year={2025},
  publisher={IEEE}
}

@inproceedings{lai2024online,
  title={Online multi-level contrastive representation distillation for cross-subject fNIRS emotion recognition},
  author={Lai, Zhili and Qing, Chunmei and Tan, Junpeng and Luo, Wanxiang and Xu, Xiangmin},
  booktitle={Proceedings of the 1st International Workshop on Brain-Computer Interfaces (BCI) for Multimedia Understanding},
  pages={29--37},
  year={2024}
}

@inproceedings{schmidt2018introducing,
  title={Introducing wesad, a multimodal dataset for wearable stress and affect detection},
  author={Schmidt, Philip and Reiss, Attila and Duerichen, Robert and Marberger, Claus and Van Laerhoven, Kristof},
  booktitle={Proceedings of the 20th ACM international conference on multimodal interaction},
  pages={400--408},
  year={2018}
}

@article{pan2023review,
  title={A review of multimodal emotion recognition from datasets, preprocessing, features, and fusion methods},
  author={Pan, Bei and Hirota, Kaoru and Jia, Zhiyang and Dai, Yaping},
  journal={Neurocomputing},
  volume={561},
  pages={126866},
  year={2023},
  publisher={Elsevier}
}

@article{udahemuka2024multimodal,
  title={Multimodal emotion recognition using visual, vocal and physiological signals: a review},
  author={Udahemuka, Gustave and Djouani, Karim and Kurien, Anish M},
  journal={Applied Sciences},
  volume={14},
  number={17},
  pages={8071},
  year={2024},
  publisher={MDPI}
}

@article{jemiolo2022datasets,
  title={Datasets for automated affect and emotion recognition from cardiovascular signals using artificial intelligence—a systematic review},
  author={Jemio{\l}o, Pawe{\l} and Storman, Dawid and Mamica, Maria and Szymkowski, Mateusz and {\.Z}abicka, Wioletta and Wojtaszek-G{\l}{\'o}wka, Magdalena and Lig{\k{e}}za, Antoni},
  journal={Sensors},
  volume={22},
  number={7},
  pages={2538},
  year={2022},
  publisher={MDPI}
}

@article{makantasis2023lab,
  title={From the lab to the wild: Affect modeling via privileged information},
  author={Makantasis, Konstantinos and Pinitas, Kosmas and Liapis, Antonios and Yannakakis, Georgios N},
  journal={IEEE Transactions on Affective Computing},
  volume={15},
  number={2},
  pages={380--392},
  year={2023},
  publisher={IEEE}
}

@article{saganowski2022emotion,
  title={Emotion recognition for everyday life using physiological signals from wearables: A systematic literature review},
  author={Saganowski, Stanis{\l}aw and Perz, Bartosz and Polak, Adam G and Kazienko, Przemys{\l}aw},
  journal={IEEE Transactions on Affective Computing},
  volume={14},
  number={3},
  pages={1876--1897},
  year={2022},
  publisher={IEEE}
}

@article{bota2019review,
  title={A review, current challenges, and future possibilities on emotion recognition using machine learning and physiological signals.},
  author={Bota, Patricia J and Wang, Chen and Fred, Ana LN and Da Silva, Hugo Pl{\'a}cido},
  journal={IEEE access},
  volume={7},
  number={99},
  pages={140990--141020},
  year={2019}
}

@article{yin2017recognition,
  title={Recognition of emotions using multimodal physiological signals and an ensemble deep learning model},
  author={Yin, Zhong and Zhao, Mengyuan and Wang, Yongxiong and Yang, Jingdong and Zhang, Jianhua},
  journal={Computer methods and programs in biomedicine},
  volume={140},
  pages={93--110},
  year={2017},
  publisher={Elsevier}
}

@inproceedings{tung2019similarity,
  title={Similarity-Preserving Knowledge Distillation},
  author={Tung, Frederick and Mori, Greg},
  booktitle={Proceedings of the IEEE/CVF International Conference on Computer Vision},
  pages={1365--1374},
  year={2019}
}

@article{huang2017like,
  title={Like What You Like: Knowledge Distill via Neuron Selectivity Transfer},
  author={Huang, Zehao and Wang, Naiyan},
  journal={arXiv preprint arXiv:1707.01219},
  year={2017}
}

\newpage

\section{Biography Section}

\begin{IEEEbiographynophoto}{Bojing Hou}
is currently a Ph.D. student in the Computational Media and Arts Thrust at The Hong Kong University of Science and Technology (Guangzhou), Guangzhou, China. Her research interests lie in multimodal affective computing, with a particular focus on reliable cross-modal knowledge transfer, uncertainty-aware multimodal learning, and weakly supervised emotion recognition.
\end{IEEEbiographynophoto}

\begin{IEEEbiographynophoto}{Ruohao LI} 
is currently a Ph.D. Candidate in the Thrust of Computational Media and Arts at the Hong Kong University of Science and Technology (Guangzhou). His research interests include Human-Robot Interaction, Context-aware Interactive System, and Immersive Technologies. He serves as a reviewer for CHI, IMWUT, ACM MM, etc.
\end{IEEEbiographynophoto}

\begin{IEEEbiographynophoto}{Yitong Zhu} 
is currently a Ph.D. student in the Thrust of Computational Media and Arts at the Hong Kong University of Science and Technology (Guangzhou). Her recent research interests include Sentiment Analysis, Human Factors Engineering and Cognitive Neuroscience. She has served as a reviewer of ACL, CVPR and ACM MM etc.
\end{IEEEbiographynophoto}

\begin{IEEEbiographynophoto}{Hongjun Liu} 
is currently a Ph.D. candidate in the Center of Data Science at New York University, advancing agentic AI and reasoning-intensive NLP through executable skill interventions, process-level rewards, and multimodal systems. She serves as a reviewer for ACL, EMNLP, ICLR, Neurips, etc.
\end{IEEEbiographynophoto}

\begin{IEEEbiographynophoto}{Dr. Luwen Yu} is currently an Assistant Professor in Computational Media and Arts at The Hong Kong University of Science and Technology (Guangzhou). She received her bachelor’s and master’s degrees in Design from Huazhong University of Science and Technology, and her PhD from the University of Leeds. Her research interests include colour science and design, colour perception and cognition, colour and emotion, visual attention, and aesthetic neuroscience, with a particular interest in multimodal psychophysiological methods and their applications in design and digital media.
\end{IEEEbiographynophoto}

\begin{IEEEbiographynophoto}{Dr. Yuyang Wang}
is an Assistant Professor in the Computational Media and Arts (CMA) Thrust at the Hong Kong University of Science and Technology (Guangzhou) Information Hub. He was with the CMA thrust from 2022 to 2024, where he held the position of Postdoctoral Research Fellow. He received a PhD in computer science from the Arts et Métiers Institute of Technology, ParisTech in France. Under the framework of the French‐German Institute for Industry of the Future, he was a visiting researcher at the Karlsruhe Institute of Technology in Germany. His research interests include affective computing,multimodal BCI and immersive human-machine interaction.
\end{IEEEbiographynophoto}

\vfill

\end{document}